\documentclass[10pt, twocolumn, twoside]{IEEEtran}
\usepackage{graphicx}
\usepackage{amsmath,amssymb} 
\usepackage{color,cite,balance}

\usepackage{url}
\usepackage{float}
\usepackage{multirow, array, longtable}
\usepackage{adjustbox}
\usepackage{xspace}

\DeclareRobustCommand{\ie}{i.e.\@\xspace}

\makeatletter
\DeclareRobustCommand{\etc}{%
	\@ifnextchar{.}%
	{etc}%
	{etc.\@\xspace}%
}
\makeatother

\definecolor{OliveGreen}{RGB}{51, 102, 51}
\definecolor{MandarinOrange}{RGB}{204, 51, 0}
\definecolor{DarkRed}{RGB}{246, 0, 0}
\definecolor{DarkBlue}{RGB}{51, 0, 153}
\newcommand{\correct}[1]{\emph{\tbf{\textcolor{DarkBlue}{#1}}}}
\newcommand{\wrong}[1]{\tbf{\textcolor{DarkRed}{#1}}}

\newcolumntype{`}{>{\global\let\currentrowstyle\relax}}
\newcolumntype{~}{>{\currentrowstyle}}
\newcolumntype{M}[1]{>{\centering\arraybackslash}m{#1}}

\newcommand{\rowstyle}[1]{\gdef\currentrowstyle{#1}#1\ignorespaces }
\newcommand{\rbf}{\rowstyle{\bfseries}}
\newcommand{\tbf}[1]{\textbf{#1}}

\newcommand{\Tst}{\rule{0pt}{2.6ex}}			
\newcommand{\Bst}{\rule[-0.9ex]{0pt}{0pt}}   
\newcommand{\TBst}{\Tst\Bst}

\newcommand{\norm}[1]{\left\lVert#1\right\rVert}
\newcommand{\gph}[2]{\includegraphics[width=#1\linewidth]{#2}}
\newcommand{\mtr}[3]{\multirow{#1}{#2\linewidth}{#3}}
\newcommand{\fmtr}[2]{\multirow{#1}{*}{#2}}

\newcommand{\Enc}{radix encoding}
\newcommand{\EncCaps}{Radix Encoding}
\newcommand{\EncS}{Radix}

\begin{document}
\title{COMIC: Towards A Compact \\ Image Captioning Model with Attention}


\author{Jia~Huei~Tan,~Chee~Seng~Chan,~\IEEEmembership{Senior~Member,~IEEE}, and~Joon~Huang~Chuah,~\IEEEmembership{Senior~Member,~IEEE}
\thanks{Manuscript received July 05, 2018; revised December 22, 2018; accepted on February 22, 2019. This research is supported by the UM Frontier Research Grant FG002-17AFR, from University of Malaya. The associate editor coordinating the review of this manuscript and approving it for publication was Dr. El Saddik, Abdulmotaleb. \textit{(Corresponding author: Chee Seng Chan)}}
\thanks{J.H. Tan and C.S. Chan are with the Center of Image and Signal Processing, Department of Artificial Intelligence, Faculty of Computer Science and Information Technology, University of Malaya, Kuala Lumpur,
50603 MALAYSIA. e-mail: \{tanjiahuei@siswa.um.edu.my; cs.chan@um.edu.my\}}
\thanks{J.H. Chuah is with the Department of Electrical Engineering, Faculty of Engineering, University of Malaya, Kuala Lumpur, 50603 MALAYSIA. e-mail: \{jhchuah@um.edu.my\}}
}

\markboth{Accepted to appear at IEEE Transaction on Multimedia}%
{IEEE Transaction on Multimedia}

\maketitle

\IEEEtitleabstractindextext{%
\begin{abstract}
	Recent works in image captioning have shown very promising raw performance. However, we realize that most of these encoder-decoder style networks with attention do not scale naturally to large vocabulary size, making them difficult to be deployed on embedded system with limited hardware resources. This is because the size of word and output embedding matrices grow proportionally with the size of vocabulary, adversely affecting the compactness of these networks. To address this limitation, this paper introduces a brand new idea in the domain of image captioning. That is, we tackle the problem of compactness of image captioning models which is hitherto unexplored. We showed that, our proposed model, named COMIC for COMpact Image Captioning, achieves comparable results in five common evaluation metrics with state-of-the-art approaches on both MS-COCO and InstaPIC-1.1M datasets despite having an embedding vocabulary size that is 39$\times$ - 99$\times$ smaller. The source code and models are available at: \url{https://github.com/jiahuei/COMIC-Compact-Image-Captioning-with-Attention}

\end{abstract}

}

\maketitle

\IEEEdisplaynontitleabstractindextext

\IEEEpeerreviewmaketitle

\section{Introduction}
\label{sec: Introduction}
Automatically generating a caption that describes an image, a problem known as image captioning, is a challenging problem where computer vision meets natural language processing. Compared to image classification and object recognition tasks, image captioning requires a higher level of scene understanding as well as language modelling. A well performing model not only has to identify the objects in the image, but also capture the semantic relationship between them, general context and the activities that they are involved in. Furthermore, the model has to map the visual representation into a fully-formed English sentence.

Given the many similarities shared between image captioning and neural translation, many recent approaches in the image captioning domain have been inspired by the advances in neural translation \cite{bahdanau2014neural,cho2014learning,sutskever2014sequence}. A common framework is to use a word embedding matrix to produce a word embedding vector to serve as the input, and a separate output projection matrix to produce a probability distribution over all the words. However we found out that when the datasets used to train the models grow larger in size, so does the vocabulary size. These huge embedding matrices in turn inflate the model, adversely affecting the compactness of the models. As a result of that, it makes them difficult to be deployed on embedded system with limited hardware resources. For example, the Recurrent Neural Network (RNN) decoder in the {\it Show, Attend and Tell} framework \cite{xu2015show} has a vocabulary size of $9,962$. The resulting model has $12.2 M$ parameters where $7.7 M$ belongs to the word embedding and output projection matrices. Even with embeddings weight sharing \cite{press2016using}, the model still has $7.3 M$ parameters where $2.6 M$ belongs to the embedding matrices. On the other hand, character-based models although compact with a small vocabulary size, suffers from poor performance. This is because character-based text sequences usually have much longer sequence lengths which exacerbates difficulties with long-range dependencies.

\begin{figure}[t]
	\begin{center} 
		\includegraphics[keepaspectratio=true, scale = 0.35]{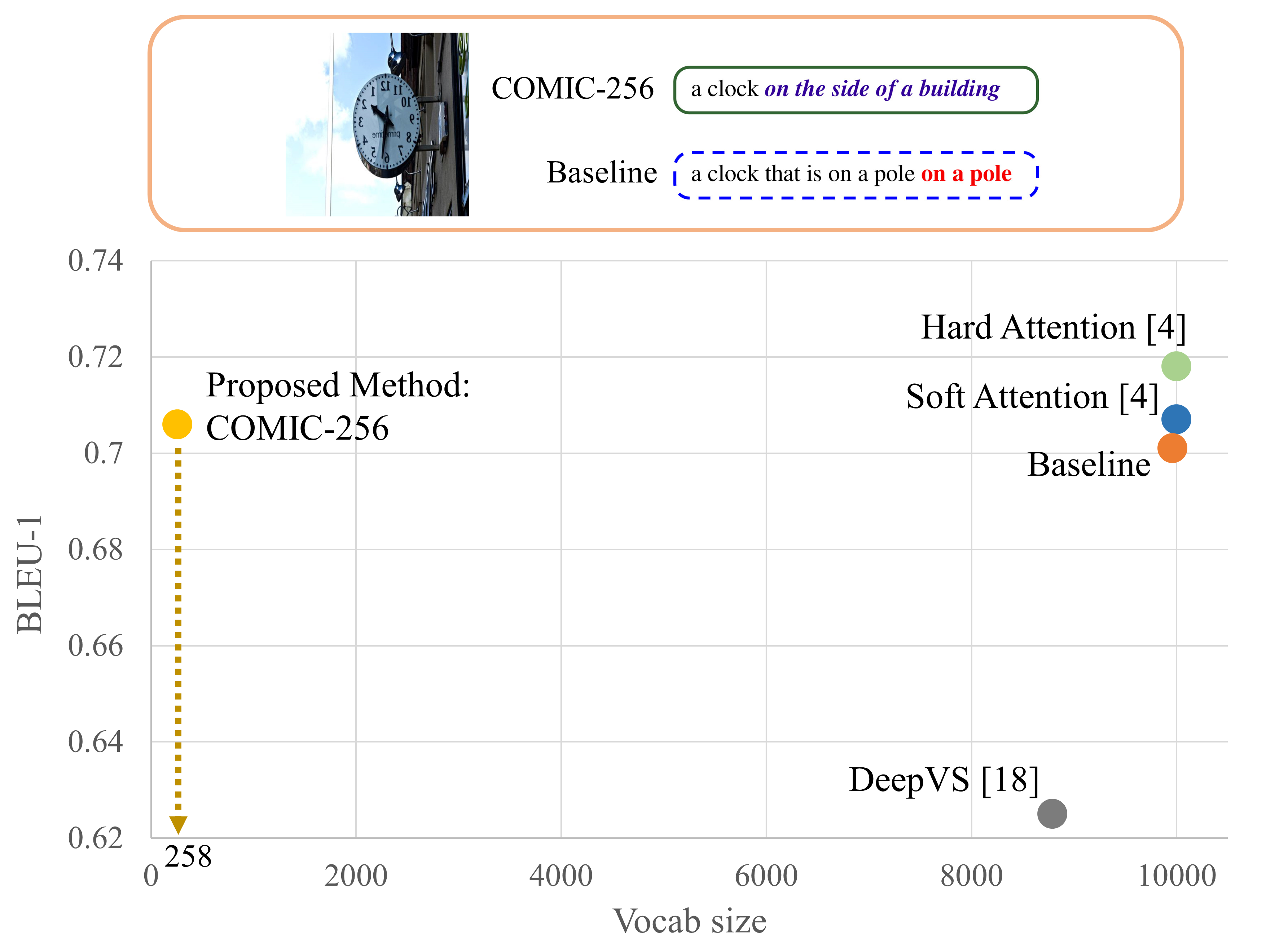}
	\end{center}
	\caption{Our proposed method (COMIC-256) is able to achieve comparable BLEU-1 score on MS-COCO (0.706) against DeepVS \cite{karpathy2015deep} (0.625); Soft-Att. \cite{xu2015show} (0.707); Hard-Att. \cite{xu2015show} (0.718) and Baseline (0.701) despite having an embedding vocabulary size that is 39$\times$ smaller.}
	\label{fig: Teaser}
	\vspace{-10pt}
\end{figure}

In this paper, our goals are to i) reduce the complexity of image captioning model without compromising the performance; and at the same time ii) improve model performance with attention module without incurring additional computational costs, paving the way for possible real-time applications deployment in resource constrained devices such as embedded/mobile devices. To achieve this goal, we present a simple yet effective framework named COMIC to reduce the model complexity in a manner that preserves the original accuracy; and at the same time increase model accuracy with attention module in a manner that preserves the model complexity.

Firstly, \EncCaps{} is employed as a pre-processing step that allows us to encode a vocabulary of size $v^d$ using $v$ symbols. The encoding scheme is designed in such a way that it can be deployed without requiring any changes to the existing image captioning models. Secondly, attention module for image captioning has become the de facto standard nowadays largely due to the success of \cite{xu2015show,you2016image,fu2016aligning, gao2017video, cho2015describing}. However, it is known that attention module usually operates on the high level Convolutional Neural Network (CNN) feature maps that come with a large channel dimension, leading to an increase in the model complexity in terms of RNN input size and their weight matrices. To combat this, we refine the feature map projection weight tying as a down-projection so that the new projected feature map has lesser channels, and thus it provides a more compact representation via attention. Finally, we adopted multi-head additive attention to take advantage of improving the effective resolution of attention module without affecting the original model complexity. With this, COMIC will have the ability to jointly attend to information from different representation subspaces at different positions. Technically, this is achieved by separating the feature map channels into groups, so that each attention head can attend to different parts of the image separately depending on the channel group. In order to prevent an increase in the computational cost due to the multi-head module, the dimensionality of each attention head is reduced by a factor of $g$, where $g$ is the number of attention heads.

In summary, the core contributions of this work are twofold. Firstly, we propose COMIC, a COMpact Image Captioning model with vastly reduced vocabulary size (up to 99$\times$ smaller) and multi-head attention module (see Section \ref{sec: Towards Compact Models}). This is the first attempt in the image captioning domain, and it opens up a new research angle in this domain. Secondly, we demonstrate the effectiveness of COMIC on two benchmark datasets: MS-COCO \cite{lin2014microsoft} and InstaPIC-1.1M \cite{chunseong2017attend} (see Section \ref{subsubsec: SOTA comparison}). We show that COMIC achieves comparable results ($\le1-2\%$ loss only) on BLEU \cite{papineni2002bleu}, METEOR \cite{banerjee2005meteor}, ROUGE-L \cite{lin2004rouge}, CIDEr \cite{vedantam2015cider} and SPICE \cite{anderson2016spice} against state-of-the-art (SOTA) methods despite having an embedding vocabulary size that is 39$\times$ - 99$\times$ smaller. We discuss the technical differences as compared to some related works in the next section.

\section{Related Works}
\label{sec: Related Works}
Our work is mostly related to the current research on image captioning and compact model. This section reviews the most relevant works on these two topics.

\tbf{Image captioning.} \cite{kiros2014multimodal} proposed a multimodal log-bilinear model to generate image captions, while  \cite{karpathy2015deep} used a Bidirectional RNN (BRNN) and Region CNN (R-CNN) to learn multimodal embedding which is then used by an RNN to generate sentences. \cite{vinyals2015show} proposed to map image features from CNN to a common word embedding space, and generating sentences using Long-Short Term Memory (LSTM) network. Their work is extended by Xu et al. \cite{xu2015show} who incorporated an attention mechanism, allowing the network to focus on salient objects. Following this, \cite{yang2016review} further extended this framework by adding a reviewer stage between the encoder and decoder.
Tan and Chan \cite{tan2016phi,TanC19} proposed a phrase LSTM model, which has two levels of LSTMs, one to model the sentence composed of phrases, and another to generate words in a phrase. \cite{fang2015captions} used multi-instance learning framework to learn 1000 visual detectors as the conditional inputs to a language model, and You et al. \cite{you2016image} enhanced the performance by learning the semantic attention on visual attributes. More examples of attribute models include \cite{wu2016value,yao2016boosting}. Park et al. \cite{chunseong2017attend} proposed to use context memory to personalise the captions for Instagram images. Wang et al. \cite{wang2016image} proposed a deep bidirectional LSTM model to harness history and future context information, and is extended by \cite{wang2018image} with the integration of multi-task learning. Dai and Lin \cite{dai2017contrastive} proposed Contrastive Learning to encourage distinctiveness of the generated captions. Although most of the aforementioned approaches achieved very promising results, all of these models do not scale naturally to large vocabulary size. Most if not all recent image captioning works focused on raw performance with the built of exotic encoder-decoder style networks with attention and placed little emphasis on reducing the computational costs of their models. In this paper, we introduce to the community a new research direction - a compact model with attention named COMIC.

\tbf{Compact model.} Building a compact model is an ongoing effort in the domain of deep learning \cite{iandola2016squeezenet,howard2017mobilenets}. In this paper, we will focus on efforts in the field of neural natural language processing, as it is closer to image captioning. There are many existing works involving the use of encoding as a pre-processing step. For instance, Nakagawa \cite{nakagawa2004chinese} proposed a hybrid method for Chinese and Japanese word segmentation, using word-level information for known words, and character-level information for unknown words. \cite{zhang2017encoding} studied numerous encoding methods for text classification in Chinese, English, Japanese and Korean. \cite{chitnis2015variable} encoded rare words using Huffman encoding into subword symbols. Similarly \cite{sennrich2015neural} proposed using Byte-Pair Encoding (BPE) to segment rare words into subword units. However, it can only be used on English or languages with Latin character. Gillick et al. \cite{gillick2016multilingual} treated the text as a sequence of variable-length UTF-8 bytes for text sequence annotation. While it does not involve natural language sequence generation, the work allows for a more compact representation of the word sequence. \cite{ke2017radical} proposed using CNN as encoder to produce radical-embeddings for Chinese and Japanese, resulting in reduced embedding vocabulary size. Li et al. \cite{li2016lightrnn} proposed building a word embedding table to factorise each word prediction into a 2-step process. The word embedding table is optimised separately using the minimum cost maximum flow (MCMF) algorithm. In our work, we show that it is possible to achieve good performance (i.e. generate a decent image caption) despite having an extremely small embedding vocabulary size.

\tbf{Summary.} Compared to regular image captioning models, COMIC has vastly fewer learnable parameters, leading to reduced requirement on GPU memory and storage. A closely related work to ours is LightRNN \cite{li2016lightrnn} but with few differences - i) COMIC requires only a single word embedding matrix (as opposed to two in LightRNN); ii) COMIC does not necessitate any changes in the model architecture (LightRNN requires a word embedding table); and iii) LightRNN is applied for language modelling only. On the other hand, our proposed method is orthogonal to compression and pruning based methods such as \cite{han2015deep,rastegari2016xnor}. Compression methods encode the trained weights of a full CNN into a smaller representation, while pruning methods are applied only after the full dense model has started the training process. In contrast, our method directly reduces the number of learnable parameters in the first place, thus producing a compact model. Moreover, \cite{han2015deep,rastegari2016xnor} are applied for image classification instead of image captioning. We believe that the aforementioned methods can be applied on top of COMIC to achieve even higher savings in terms of storage and parameters. 


\section{Overall Architecture}
\label{sec: Overall Architecture}
Following recent works, we formulate the image captioning task as a translation problem, where a probabilistic model is used to ``translate'' an image with fixed-size representation into a fully-formed English sentence. As such, we adopt a modified version of {\it Show, Attend and Tell} \cite{xu2015show} framework as our model architecture, since it provides good performance on the image captioning task. This model will also serve as the baseline for our experiments. For clarification, we will refer to output projection and output embedding; embedding dimension and word size interchangeably. All the model size calculations in this paper include only the decoder and attention module (the encoder, \ie{} CNN is excluded).

Suppose $\left\{S_0,\: \cdots\:, \:S_L\right\}$ is a sequence of words, our model directly maximises the probability of the correct description given an image $I$ using the following formulation:
\small
\begin{equation}\label{eq: log-likelihood}
	\log{p}\left( S\,|\,I \right)=\sum_{t\,=\,0}^L\log{p}\left( S_t\,|\,I,\; S_{0\::\:t-1}, \;c_t \right)
\end{equation}
\normalsize
\noindent where $p\left( S_t\,|\,I,\; S_{0\::\:t-1},\; c_t \right)$ is the probability of generating a word given an image $I$, previous words $S_{0\::\:t-1}$, and context vector $c_t$. 

Although in principle any RNNs can be used, LSTM cell \cite{hochreiter1997long} (with forget bias) is chosen as it has shown SOTA performance on sequential tasks such as translation \cite{britz2017massive} and image captioning. For a LSTM network with $n$ units, we initialise the hidden state of LSTM with image embedding vector through a pre-activation weight layer with layer normalisation (LN) \cite{ba2016layer}:
\begin{equation}\label{eq: LSTM init}
	h_{t=-1} = W_I\,\tanh\left( LN\left(I_{embed}\right)\right)
\end{equation}
\noindent where $W_I \in\mathbb{R}\:^{n \times z}$ is a weight matrix and $LN\left(\cdot\right)$ is the LN function.

The attention function used in this paper is the ``soft-attention'' introduced by \cite{bahdanau2014neural} and used in \cite{xu2015show}, where a multilayer perceptron (MLP) with a single hidden layer is employed to calculate the attention weights on a particular feature map. The context vector $c_t$ is then concatenated with previous predicted word embedding to serve as input to the LSTM. Finally, a probability distribution over the vocabulary is produced from the hidden state $h_t$:

\begin{equation}\label{eq: LSTM output}
	p\,_t = Softmax \left( E_o\,h_t \right)
\end{equation}
\begin{equation}\label{eq: LSTM}
	h_t \:,\: m_t = LSTM\left(x_t \:,\: h_{t-1} \:,\: m_{t-1}\right)
\end{equation}
\begin{equation}\label{eq: LSTM input}
	x\,_t = \left[ E_w\:S_{t-1},\; c_t \right]
\end{equation}
\begin{equation}\label{eq: Context}
	c_t = \sum_{j}^{|F|} \left( \alpha_{tj} \odot f_j \right)
\end{equation}
\begin{equation}\label{eq: Attention}
	\alpha_{tj} = \frac{\exp\left( MLP\left(f_j\,,h_{t-1}\right) / \epsilon\right)}{\sum_{j}^{|F|} \exp\left( MLP\left(f_j\,,h_{t-1}\right) / \epsilon\right)}
\end{equation}
\small
\begin{equation}\label{eq: Attention MLP}
	MLP\left(f_j\,,h_{t-1}\right) = W_{M2} \tanh\left( LN\left(W_{M0}\:f_j + W_{M1}\:h_{t-1}\right)\right)
\end{equation}
\normalsize
\noindent where $E_w \in\mathbb{R}\,^{m \times v}$ and $E_o \in\mathbb{R}\,^{v \times n}$ are input and output embedding matrices respectively; $W_{M0} \in\mathbb{R}\,^{k \times r}$, $W_{M1} \in\mathbb{R}\,^{k \times n}$, $W_{M2} \in\mathbb{R}\,^{1 \times k}$ are weight matrices; and $\left[ \,,\, \right]$ is the concatenation operator. $p_t$ is the probability distribution over the vocabulary; $m_t$ is the memory state; $x_t$ is the current input to LSTM; $c_t \in\mathbb{R}\,^q$ is the context vector; $f \in\mathbb{R}\,^{|F| \times r}$ is the feature map and $f_j \in\mathbb{R}\,^{r}$ is the vector extracted from location $j$\,; $\alpha_{tj}$ is the attention weight at time step $t$ and location $j$; $S_{t-1} \in\mathbb{R}\,^m$ is the one-hot vector of previous word; $\epsilon$ is the softmax temperature.


\section{Towards Compact Image Captioning Model}
\label{sec: Towards Compact Models}
This paper introduces COMIC, a simple yet effective framework that consists of \Enc{}, feature map projection weight tying and multi-head additive attention that work together to built a compact image captioning model with attention without affecting the original accuracy.

\subsection{\EncCaps{}}
\label{subsec: Proposed Encoding}
\begin{figure}[t]
	\begin{center} 
		\includegraphics[keepaspectratio=true, scale = 0.35]{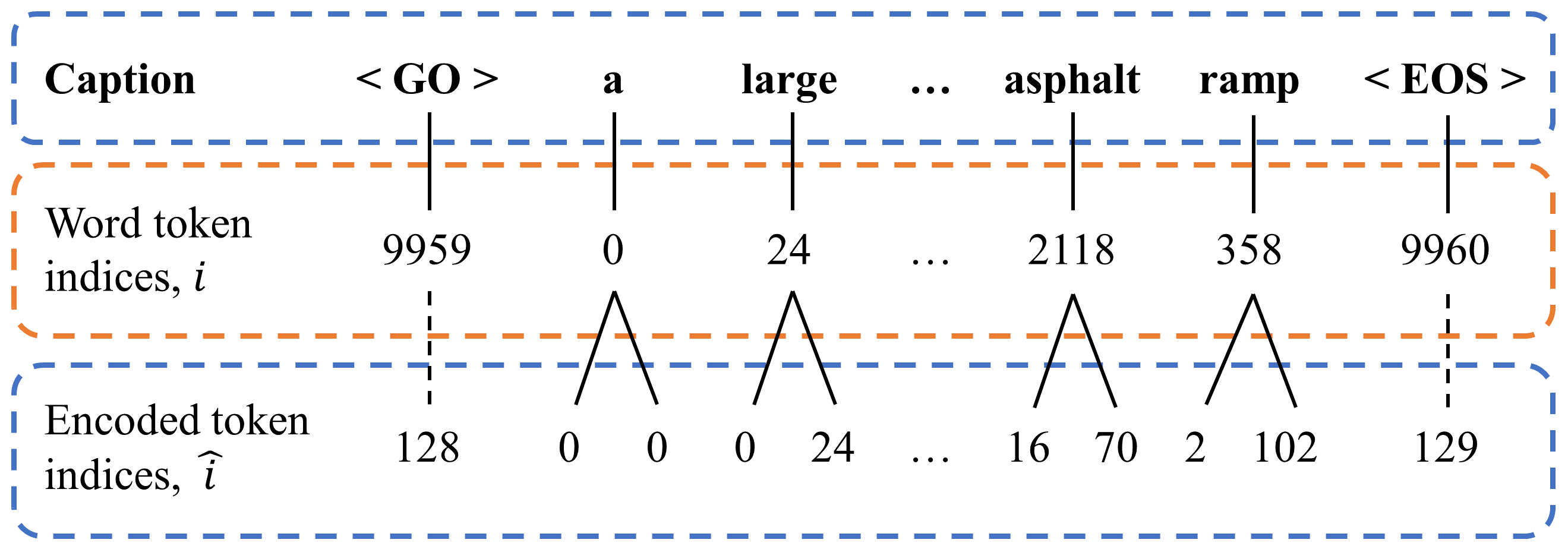}
	\end{center}
	\caption{Example of caption, original and encoded token indices when using \Enc{} with base-128}
	\label{fig: Encoding}
\end{figure}

The idea of the \Enc{} is to transform the word indices to a higher base, splitting every word token into $d$ tokens where $d \geq 2$. Although it is possible to achieve reduction in the vocabulary size using BPE \cite{sennrich2015neural}, it can only be used on English and other languages using Latin characters. On the other hand, \Enc{} can in theory be used on all languages including Chinese, Japanese, Korean \etc. For example in Fig. \ref{fig: Encoding}, with a base of $v = 128$, the word token ``a'' with an index of $i = 0$ will be encoded using two tokens of $\hat{i_0} = 0$ and $\hat{i_1} = 0$; while the word token ``asphalt'' with an index of $i = 2118$ will be encoded using two tokens of $\hat{i_0} = 16$ and $\hat{i_1} = 70$. 
We also define two special tokens where $\langle$GO$\rangle$ marks the start of a sequence and $\langle$EOS$\rangle$ marks the end of the sequence. For easy decoding, the special tokens are represented using only one token. $\langle$GO$\rangle$ is assigned with an index of $\hat{i} = v$ (one-hot vector $e^{(v)}$) and $\langle$EOS$\rangle$ is assigned with an index of $\hat{i} = v + 1$ (one-hot vector $e^{(v+1)}$). This enables \Enc{} to be used without any modification to the existing model architectures. To generate a sequence, one simply run inference using beam search as usual and apply post-processing on the output tokens. The post-processing can be done by either converting the encoded index $\hat{i_0}, \hat{i_1}$ back to base-10 index $i$, or by constructing a decoding tree dictionary.

With this encoding scheme, we managed to re-represent the original corpus vocabulary $V_o$ of size $v^d$ using an encoded embedding vocabulary $V_e$ of size $v$. This leads to a huge reduction in the model complexity. For example, the popular MS-COCO dataset often yields a vocabulary of $8,000$ to $10,000$ words while the InstaPIC-1.1M dataset yields around $22,000$ to $40,000$ words. With the proposed \Enc{}, $V_e$ can be set to a much lower size such as $v = 128$ or $v = 256$, a reduction of almost $39\times$ in the MS-COCO dataset and $99\times$ in the InstaPIC-1.1M dataset. Results are given in Section \ref{subsubsec: Tokenisation and encoding}, Table \ref{table: Tokenisation performance}.

\subsection{Feature Map Projection Weight Tying}
\label{subsec: Feature Map Projection}

For most of the image captioning models with visual attention \cite{xu2015show,you2016image,fu2016aligning, gao2017video, cho2015describing}, the attention function operates on higher level feature map of the CNN in order for the context vector $c_t$ to capture higher level representation. Such feature maps usually comes with a large channel dimension, such as $r = 512$ for VGG-16 and $r = 832$ for GoogLeNet. This in turn increases the RNN's input size and their weight matrices. To combat this issue, we propose a down-projection algorithm on the feature map such that the projected map has lesser channels, given by
\begin{equation}\label{eq: FM down project untied}
	c_t = \sum_{j}^{|F|} \left( \alpha_{tj} \odot W_{f}\:f_j \right)
\end{equation}
\noindent where $W_{f} \in\mathbb{R}\,^{q \times r}$ is a weight matrix, $q$ is the number of channels of the projected feature map and $q \ll r$.  As shown in Table \ref{table: Attention} (untied), a small projection size such that $q \ll r$ can reduce the model complexity, and at the same time it provides extra representation power to the language model.

However, still, the extra projection layer will naturally incur additional computational cost. To further alleviate this complexity issue, we introduce weight sharing on the feature map projection $W_{f}$ and attention MLP weights $W_{M0}$\,, given by
\begin{equation}\label{eq: FM down project tied}
	c_t = \sum_{j}^{|F|} \left( \alpha_{tj} \odot W_{M0}\:f_j \right)
\end{equation}
\noindent With this, the projected feature map $W_{M0}\,f_j$ can be calculated in advanced and share with the attention MLP, and so the visual attention module can be introduced in COMIC without incurring extra computational cost as to conventional approaches. Table \ref{table: Attention} shows that the attention module is put forward in a lower computational cost without compromising the accuracy. 

\subsection{Multi-Head Additive Attention}
\label{subsec: Multi-Head Attention}

Multi-head attention \cite{vaswani2017attention} separates the feature map channels into groups, where each attention head can attend to different areas of the image separately depending on the channel group. In other words, each location of each channel group is assigned an attention weight with a MLP. This approach is opposed to the regular single-head attention which applies attention weights equally across all of the feature map channels, leading to averaging of contextual information from multiple regions.

Technically, for single-head attention, we use a single MLP with hidden size $k$ and obtain a $q$-dimensional context vector $c_t$. For multi-head attention with $g$ heads, we use $g$ separately learned MLPs with hidden size $k / g$, with each head produces a $q / g$-dimensional output vector. The output vectors are then concatenated to produce the final context vector $c_t$ with $q$-dimensions. Due to the dimension reduction of each head, the total computational cost will be the same as to the single-head attention with full dimensionality. 

To take advantage of this, we apply the multi-head dot-product attention\cite{vaswani2017attention}, together with additive attention (MLP) and image captioning to increase the effective resolution of the attention module via an ensemble of attention modules. In practise, we combine the separate MLPs into a single MLP to maximise parallelism. Experiments on MS-COCO dataset (Table \ref{table: Attention}) show that multi-head can improve the original accuracy with same model complexity. This is the first time multi-head additive attention is used in image captioning\footnote{\cite{kaiser2017one} employed multi-head dot-product attention, however no result on image captioning is provided.}. 

\begin{table*}[ht]
	\centering
	\caption{Comparison of models with different tokenisation and encoding schemes on MS-COCO}
	\label{table: Tokenisation performance}
		\begin{tabular}{`l|~c|~c ~c ~c ~c ~c ~c ~c ~c}
			\hline \TBst
			Tokens					& \# params.	& B-1   & B-2   & B-3   & B-4   & M     & R     & C     & S     \\
			
			\hline \Tst
			
			Character      			& 4.5 M			& 0.670 & 0.498 & 0.364 & 0.266 & 0.220 & 0.495 & 0.770 & 0.149 \\ \Bst
			Word      				& 12.2 M \rbf	& 0.704 & 0.533 & 0.397 & 0.295 & 0.235 & 0.517 & 0.880 & 0.165 \\
			
			\hline \Tst
			
			\EncS{}, base-64		& 4.5 M			& 0.693 & 0.517 & 0.380 & 0.280 & 0.229 & 0.507 & 0.824 & 0.155 \\ \Bst
			\EncS{}, base-128		& 4.6 M \rbf	& 0.694 & 0.522 & 0.386 & 0.287 & 0.233 & 0.509 & 0.848 & 0.159 \\
			\hline
		\end{tabular}
\end{table*}

\section{Experiments and Discussion}
\label{sec: Experiments}
\subsection{Model Details}
\label{subsec: Model Details}

The LSTM model is trained in an unrolled form to predict each word of the sentence after it has seen the image, the current context vector and all the preceding words, as given by $p\left( S_t\;|\;I,\; S_{0\::\:t-1},\; c_t \right)$. As usual, each word is represented as one-hot vector $S_t$ of dimension equal to the size of the dictionary.
The training is performed by minimising the loss w.r.t. all the parameters except the image model. To tackle overfitting, we employed dropout at the input and output of the LSTM. Our loss function is the sum of the negative log likelihood of the correct word at each time step, doubly stochastic attention regularisation as employed in \cite{xu2015show} and L2 weight loss as given below:
\small
\begin{equation}\label{eq: objective function}
	L \left(\, I,\: S \,\right) = - \sum_{t}^{L} \: \log p_{\,t} \left( S_{\,t} \right) + \sum_{j}^{|F|} \left( 1 - \sum_{t}^{L} \alpha_{tj} \right)^2 + \lambda\cdot\norm{ \theta}^2_2
\end{equation}
\normalsize

Unless stated otherwise, all the models used in our experiments have the following basic configurations. All models are implemented using TensorFlow. The image model used in our work is GoogLeNet (InceptionV1) with batch normalisation \cite{szegedy2015going,ioffe2015batch} pre-trained on ImageNet. The input images are resized to $256 \times 256$, and randomly flipped and cropped to $224 \times 224$ before being fed to the CNN. The image embedding size is $z = 1024$. The attention function operates on the ``Mixed-4f'' map $f \in\mathbb{R}\:^{196 \times 832}$, with MLP size of $k = 512$. The projected feature map for untied models in Table \ref{table: Attention} have $q = 512$ channels. The LSTM network consists of a single layer with hidden state size of $n = 512$. The word size is set to $m = 256$ dimensions.
The optimiser used for training is Adam \cite{kingma2014adam}, with batch size of $32$.

The initial learning rate is set to $1 \times 10^{-3}$, and is halved every $4$ epochs until a minimum of $2 \times 10^{-4}$. All models are trained for $20$ epochs. The input and output dropout rates for LSTM are both set to $0.35$. Weight decay rate is set to $\lambda = 1 \times 10^{-5}$. All trainable parameters are initialised randomly using Xavier initialisation \cite{glorot2010understanding}. For inference, we used beam search in order to better approximate $S = \arg \max_{S\,\prime} \: p(S^{\:\prime} \,|\, I)$. We use beam size $b = 3$ with no length normalisation for all experiments unless noted otherwise. All hyperparameters are chosen based on educated guesses due to limited computational resources.

\subsection{Experiment Setup}
\label{subsec: Experiment Setup}
We conducted our experiments on two public English captioning datasets, namely MS-COCO \cite{lin2014microsoft} and InstaPIC-1.1M \cite{chunseong2017attend}. MS-COCO dataset contains $123,287$ images and each image is given at least $5$ captions by different AMT workers. We use the publicly available split\footnote{\label{footnote: karpathy}http://cs.stanford.edu/people/karpathy/deepimagesent/} in the work of \cite{karpathy2015deep}, which use $5,000$ images for validation, and another $5,000$ for testing. InstaPIC dataset contains $648,761$ images for training, and $5,000$ images for testing. Each Instagram image is paired with one user caption. This dataset is challenging, as its captions are natural posts with varying formats. Following \cite{dai2017contrastive}, we reserved $2,000$ images randomly from the training set for validation.

All the scores are obtained using the publicly available MS-COCO evaluation toolkit\footnote{https://github.com/peteanderson80/coco-caption} , which computes BLEU \cite{papineni2002bleu}, METEOR \cite{banerjee2005meteor}, ROUGE-L \cite{lin2004rouge}, CIDEr \cite{vedantam2015cider} and SPICE \cite{anderson2016spice}. For sake of brevity, we label BLEU-1 to BLEU-4 as B-1 to B-4, and METEOR, ROUGE-L, CIDEr, SPICE as M, R, C, S respectively. For MS-COCO, we use the publicly available tokenised captions\textsuperscript{\ref{footnote: karpathy}} \cite{karpathy2015deep}, filtering out words that occur less than $5$ times and truncating sentences longer than $20$ words. For InstaPIC, we use the publicly available tokenisation script\footnote{\label{footnote: attend2u}https://github.com/cesc-park/attend2u}, and select $25,595$ most frequently used words as our vocabulary. We also truncate captions longer than $18$ words.

\subsection{Ablation Study}
\label{subsec: Results and Comparison}

\subsubsection{Tokenisation and encoding}
\label{subsubsec: Tokenisation and encoding}

\begin{table*}[t]
	\centering
	\caption{Comparison of models with different attention configurations on MS-COCO}
	\label{table: Attention}
	\begin{adjustbox}{max width=0.9\linewidth}
		\begin{tabular}{`l|~c|~c|~c|~c ~c ~c ~c ~c ~c ~c ~c}
			\hline \TBst
			Approach							& Projection		& \# params.	& Att. heads	& B-1   & B-2   & B-3   & B-4   & M     & R     & C     & S     \\
			\hline \Tst
			
			\mtr{9}{0.08}{Word} 				& \fmtr{3}{None}	& \fmtr{3}{9.8 M} & 1			& 0.703 & 0.532 & 0.396 & 0.295 & 0.236 & 0.517 &\tbf{0.886}&\tbf{0.166}\\
			\null 								&					&				& 4				& 0.700 & 0.530 & 0.396 & 0.297 & 0.237 & 0.518 & 0.883 & 0.165 \\ \Bst
			\null  								&					&				& \tbf{8}				
			& \tbf{0.704} & \tbf{0.534} & \tbf{0.398} & \tbf{0.298} & \tbf{0.238} & \tbf{0.518} & 0.885 & \tbf{0.166} \\
			\cline{2-12} \Tst
			\null								& \fmtr{3}{Untied}	& \fmtr{3}{9.6 M} & 1			& 0.705 & 0.536 & 0.402 & 0.302 & 0.239 & 0.522 & 0.897 & 0.168 \\
			\null 								&					&				& 4				& 0.708 & 0.541 & 0.408 & 0.308 & 0.241 & 0.524 & 0.911 & 0.171 \\ \Bst
			\rbf  								&					&				& 8				& 0.714 & 0.545 & 0.410 & 0.308 & 0.242 & 0.524 & 0.917 & 0.172 \\
			\cline{2-12} \Tst
			\null								& \fmtr{3}{Tied}	& \fmtr{3}{9.2 M} & 1			& 0.704 & 0.532 & 0.396 & 0.296 & 0.237 & 0.517 & 0.894 & 0.167 \\
			\null 								&					&				& 4				& 0.711 & 0.545 & 0.410 & 0.307 & 0.242 & 0.526 & 0.921 & 0.171 \\ \Bst
			\rbf  								&					&				& 8				& 0.713 & 0.546 & 0.411 & 0.309 & 0.243 & 0.526 & 0.927 & 0.172 \\
			
			\hline \Tst
			
			\mtr{9}{0.08}{\EncS{}, base-128}	& \fmtr{3}{None}	& \fmtr{3}{4.2 M} & 1			& 0.690 & 0.518 & 0.382 & 0.282 & 0.230 & 0.507 & 0.837 & 0.159 \\
			\rbf 								&					&				& 4				& 0.695 & 0.522 & 0.385 & 0.285 & 0.231 & 0.512 & 0.840 & 0.160 \\ \Bst
			\null 								&					&				& 8		& 0.692 & 0.521 &\tbf{0.385}&\tbf{0.285}&\tbf{0.231}& 0.509 & 0.839 & 0.158 \\
			\cline{2-12} \Tst
			\null 								& \fmtr{3}{Untied}	& \fmtr{3}{3.9 M} & 1 			& 0.692 & 0.520 & 0.385 & 0.286 & 0.232 & 0.511 & 0.845 & 0.161 \\
			\rbf  								&					&				& 4				& 0.700 & 0.531 & 0.395 & 0.294 & 0.234 & 0.515 & 0.864 & 0.164 \\ \Bst
			\null  								&					&				& 8				& 0.697 & 0.525 & 0.389 & 0.289 & 0.233 & 0.513 & 0.861 & 0.162 \\
			\cline{2-12} \Tst
			\null 								& \fmtr{3}{Tied}	& \fmtr{3}{3.5 M} & 1 			& 0.692 & 0.521 & 0.387 & 0.287 & 0.230 & 0.511 & 0.840 & 0.159 \\
			\null  								&					&				& 4				& 0.696 & 0.525 & 0.390 & 0.291 & 0.232 & 0.514 & 0.852 & 0.159 \\ \Bst
			\rbf  								&					&				& 8				& 0.700 & 0.529 & 0.394 & 0.294 & 0.234 & 0.514 & 0.871 & 0.165 \\
			\hline
		\end{tabular}
	\end{adjustbox}
\end{table*}

In this section, we examine the effect of the introduction of Radix Encoding scheme. From Table \ref{table: Tokenisation performance}, it can be seen that the regular word-based model performed the best. This is followed by \EncS{} models using base-128, base-64 and finally the character-based model. The result can be attributed to the much shorter sentence length when using word tokens, which alleviates long-term dependency learning issues. Also, this performance degradation is an expected trade off of parameter reduction and we believe the result is still comparable. For instance, we can notice the performance gap between word and \Enc{} model is moderate ($2.3\%$ in average), while the number of parameters reduced drastically (by $62\%$). This is almost one-third of the original amount which is comparable to the character model, yet at the same time it obtains better performance than the character model. This shows that \Enc~is able to reduce the complexity of image captioning models without affecting much on the original accuracy.

\subsubsection{Attention module}
\label{subsubsec: Attention}

In this section, we investigate the effect of different attention configurations, by varying the number of attention heads with and without the projection weight tying. The models used are as described in Section \ref{subsec: Model Details}, but with word size set to $m = 64$. Table \ref{table: Attention} shows that it is possible to introduce visual attention module in a more compact way without compromising the original accuracy. For instance, when the feature map projection is employed (\ie tied) in \EncS{}, base-128, we found that even with lesser parameters, having the extra projection layer contributes a slight improvement in the overall performance. This is more obvious when the multi-head attention is applied. This phenomena is also spotted in the regular word-based model. Without the projection, multi-head attention often provide little to no benefit compared to regular single-head attention. This can be attributed to the extra projection provides the model ability to group channels that are relevant to each attention head together, forming contiguous groups.

From our further investigation on the type of projection, we notice that there are two opposite trends. When using single-head attention, the tied models generally performed slightly worse than the untied counterparts. This shows that the benefit obtained by the extra projection is counteracted by the reduction in the parameter count. On the other hand, the tied models generally performed better than untied counterparts when using the multi-head attention, despite having lesser parameters. This can be understood as the tied projection layer receives extra gradient information via weight sharing from both the multi-attention module and RNN, while training.

In terms of the inclusion of multi-head additive attention, we can notice that compared to a single head, models using $8$ heads yields improvements of up to $+4\%$ on CIDEr score. Furthermore, as shown in Table \ref{table: Attention}, the other metric scores also improve across the board as the number of attention heads increases, when the tied projection is used. This is consistent with the findings of \cite{vaswani2017attention}, where their performance on the WMT 2014 English-to-German translation task improves as the number of heads increases (up to $16$ heads). Note that, as aforementioned in Section \ref{subsec: Multi-Head Attention} this overall performance improvement is essentially free as each individual head operate on a reduced dimension compared to the single-head.

\begin{table*}[t]
	\centering
	\caption{Comparison with baseline and SOTA methods on MS-COCO. v2 model is trained on a different setting where f = fine-tune model and s = self-critical sequence training (SCST), please refer to our github page for more details.}
	\label{table: SOTA Performance on MS-COCO}
	\begin{adjustbox}{max width=0.9\linewidth}
		\begin{tabular}{`l|~c|~c ~c ~c ~c ~c ~c ~c ~c}
			\hline \TBst
			Approaches						        & Vocab size	& B-1   & B-2   & B-3   & B-4   & M     & R     & C     & S       \\
			
			\hline \Tst
			
			DeepVS \cite{karpathy2015deep}			& 8,791			& 0.625 & 0.450 & 0.321 & 0.230 & 0.195 & -     & 0.660 & -       \\
			Google NIC \cite{vinyals2015show}		& -				& 0.666 & 0.461 & 0.329 & 0.246 & -     & -     & -     & -       \\
			Soft-Attention \cite{xu2015show}		& 10,000		& 0.707 & 0.492 & 0.344 & 0.243 & 0.239 & -    & -     & -       \\
			Hard-Attention \cite{xu2015show}		& 10,000		& 0.718 & 0.504 & 0.357 & 0.250 & 0.230 & -    & -     & -       \\
			Review Net \cite{yang2016review}	    & 9,520			& -     & -     & -     & 0.290 & 0.237 & -     & 0.886 & -       \\
			ATT \cite{you2016image}					& -				& 0.709 & 0.537 & 0.402 & 0.304 & 0.243 & -     & -     & -       \\
			ACVT \cite{wu2016value}					& 8,791			& 0.740  & 0.560  & 0.420  & 0.310  & 0.260  & -     & 0.940  & -       \\
			SCA-CNN (VGG-19) \cite{chen2017sca} 	& - 			& 0.705 & 0.533 & 0.397 & 0.298 & 0.242 & -     & -     & - \\ \Bst
			Deep-RL \cite{ren2017deep} 				& - 			& 0.713 & 0.539 & 0.403 & 0.304 & 0.251 & 0.525 & 0.937 & - \\
			
			\hline \hline \Tst
			Baseline  (12.2 M params)    			&\fmtr{3}{9,962}&     0.701 &     0.531	& \tbf{0.396} & 0.296	& 0.238	& 0.518	& 0.885 & 0.167   \\ \Bst
			Baseline-v2  (12.7 M params)    			&  &     0.716 &     -	&- & 0.311	& -	& -	& 0.937 & 0.174   \\  \Bst
			Baseline-8  (12.2 M params)				&				&     0.703 &     0.532 &\tbf{0.396}& 0.294 & 0.235 & 0.517 & 0.883 & 0.166  \\ 
			\hline \Tst
			Baseline-SC  (3.9 M params)    			&\fmtr{2}{9,962}&     0.696 &     0.525 &     0.386 & 0.284 & 0.230 & 0.510 & 0.839 & 0.158   \\ \Bst
			Baseline-8-SC  (3.9 M params)			&				&     0.698 &     0.524 &     0.387 & 0.286 & 0.231 & 0.511 & 0.855 & 0.160  \\ 
			\hline \Tst
			COMIC-128  (3.9 M params)     			& 130			&     0.700 &     0.529 &     0.393 & 0.294 & 0.236 & 0.515 & 0.875 & 0.164   \\ \Bst
			COMIC-256  (4.0 M params)     			& 258			&0.706&\tbf{0.534}&     0.395	& 0.292	& 0.237	& 0.517	& 0.881	& 0.164   \\
			COMIC-256v2  (4.3 M params)     			& 258			&0.713&-&     -	& 0.308	& -	& -	& 0.994	& 0.176   \\
			COMIC-256v2$_f$  (4.3 M params)     			& 258			&0.729&-&     -	& 0.328	& -	& -	& 1.001	& 0.185   \\
			COMIC-256v2$_s$  (4.3 M params)     			& 258			&\tbf{0.753}&-&     -	& \tbf{0.344}	& -	& -	& \tbf{1.050} 	& \tbf{0.190}   \\

			\hline
		\end{tabular}
	\end{adjustbox}
\end{table*}

\begin{table*}[t]
	\centering
	\caption{Comparison with baseline and SOTA methods on InstaPIC-1.1M. Methods with [$^\ast$] are extracted from \cite{chunseong2017attend}}
	\label{table: SOTA Performance on InstaPIC}
	\begin{adjustbox}{max width=0.9\linewidth}
		\begin{tabular}{`l|~c|~c ~c ~c ~c ~c ~c ~c ~c}
			\hline \TBst
			Approaches						        & Vocab size	& B-1   & B-2   & B-3   & B-4   & M     & R     & C     & S       \\
			
			\hline \Tst
			
			Google NIC$^\ast$ \cite{vinyals2015show} &\fmtr{3}{40,000}& 0.055 & 0.019 & 0.007 & 0.003 & 0.038 & 0.081 & 0.004 & -       \\
			Show, Attend and Tell$^\ast$ \cite{xu2015show}	& 		& 0.106 & 0.015 & 0.000 & 0.000 & 0.026 & 0.140 & 0.049 & -       \\
			CSMN \cite{chunseong2017attend}			& 				& 0.079 & 0.032 & 0.015 & 0.008 & 0.037 & 0.120 & 0.133 & -       \\ \Bst
			AACL \cite{dai2017contrastive}			& 22,886		& 0.072 & 0.028 & 0.013 & 0.006 & 0.032 & 0.101 & 0.144 & -       \\
			
			\hline \hline \Tst 
			
			Baseline  (24.0 M params)    			&\fmtr{2}{25,598}&    0.053 &     0.022 &     0.011 &\tbf{0.006}&     0.028 &     0.086 &     0.117 &     0.020  \\ \Bst
			Baseline-8  (24.0 M params) 			&				&     0.053 &     0.022 &     0.011 &\tbf{0.006}&     0.028 &\tbf{0.087}&\tbf{0.122}&\tbf{0.025} \\
			
			\hline \Tst 
			
			Baseline-SI  (4.2 M params)    			&\fmtr{2}{25,598}&    0.040 &     0.017 &     0.008 &     0.003 &     0.021 &     0.076 &     0.095 &     0.008 \\ \Bst
			Baseline-8-SI  (4.2 M params) 			&				&     0.039 &     0.015 &     0.007 &     0.003 &     0.021 &     0.079 &     0.087 &     0.005
			 \\
			
			\hline \Tst 
			
			COMIC-160  (4.0 M params)     			& 162			&     0.059 &     0.024 &     0.011 &     0.004 &     0.027 &     0.077 &     0.100 &     0.022  \\ \Bst 
			COMIC-256  (4.0 M params)     			& 258			&\tbf{0.065}&\tbf{0.026}&\tbf{0.012}&     0.005 &\tbf{0.030}&     0.083 &     0.105 &     0.021  \\
			\hline
		\end{tabular}
	\end{adjustbox}
\end{table*}

\begin{table}[t]
	\centering
	\caption{Caption statistics: uniqueness and length}
	\label{table: Caption statistics}
	\begin{adjustbox}{max width=0.8\linewidth}
		\begin{tabular}{`l|`l|~c|~c}
			\hline \TBst
			Dataset				& Model 	& Unique captions (\%)		& Average length (words)	\\
			\hline \Tst
			\mtr{3}{0.15}{MS-COCO}      	& Baseline			& 41.70			& 9.0		\\
			\null							& Baseline-8		& 40.46			& 8.9		\\ \Bst
			\null							& COMIC-256			& \rbf 43.20	& 9.2		\\
			\hline \Tst
			\mtr{3}{0.15}{InstaPIC}      	& Baseline			& 8.12			& 3.6		\\
			\null							& Baseline-8		& 6.24			& 3.7		\\ \Bst
			\null							& COMIC-256			& \rbf 13.86	& 4.5		\\
			
			\hline
		\end{tabular}
	\end{adjustbox}
	\vspace{-5pt}
\end{table}

\begin{figure*}[ht]
	\centering
	\gph{0.9}{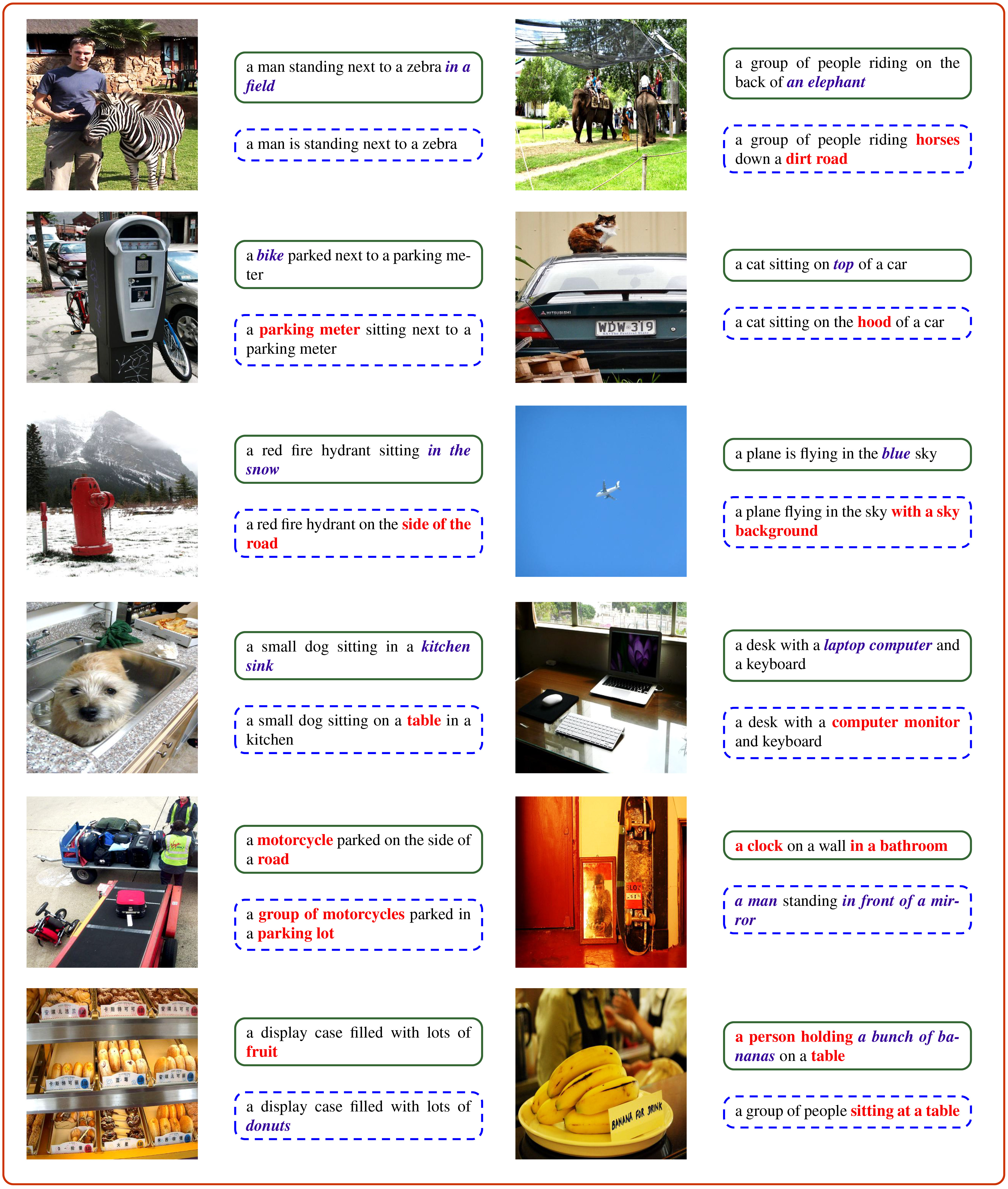}
	\caption{Captions generated by COMIC-256 and baseline on MS-COCO dataset: We can see that COMIC-256 model (solid green line) outperforms baseline method (dashed blue line) in most cases. Accurate descriptions are indicated by blue with bold and italic text, inaccurate descriptions are indicated by red with bold text. Best viewed in colour.}
	\label{fig: captions MSCOCO}
\end{figure*}

\begin{figure*}[ht!]
	\centering
	\gph{0.9}{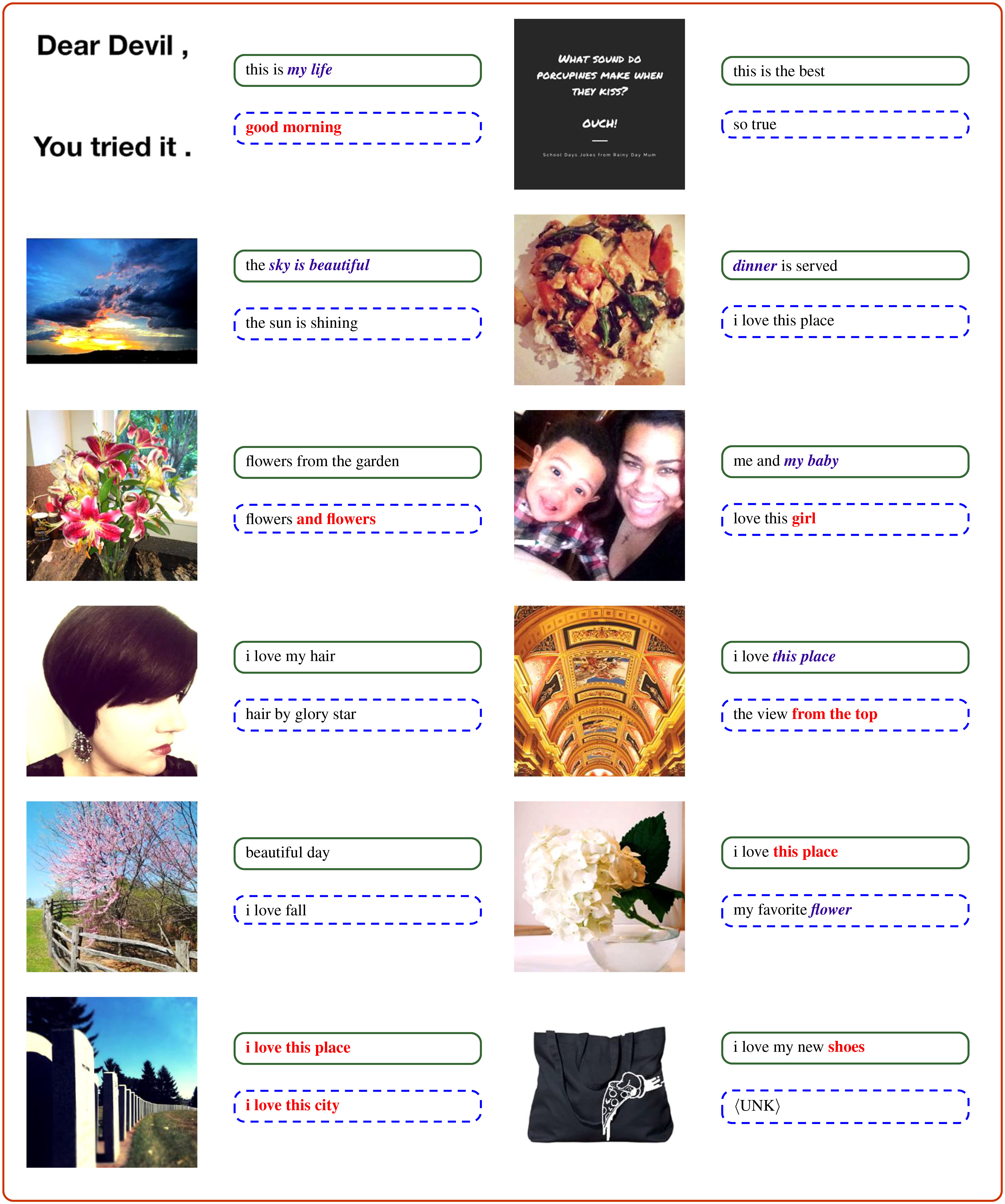}
	\caption{Captions generated by COMIC-256 and baseline on InstaPIC-1.1M dataset: We can see that COMIC-256 model (solid green line) outperforms baseline method (dashed blue line) in most cases. Accurate descriptions are indicated by blue with bold and italic text, inaccurate descriptions are indicated by red with bold text. Best viewed in colour.}
	\label{fig: captions InstaPIC}
\end{figure*}

\begin{figure*}[ht]
	\scriptsize
	\centering
	\begin{adjustbox}{max width=\linewidth}
		\begin{tabular}{M{0.15\linewidth} M{0.85\linewidth}}
			\gph{1.0}{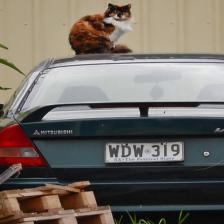}	&
			\gph{1.0}{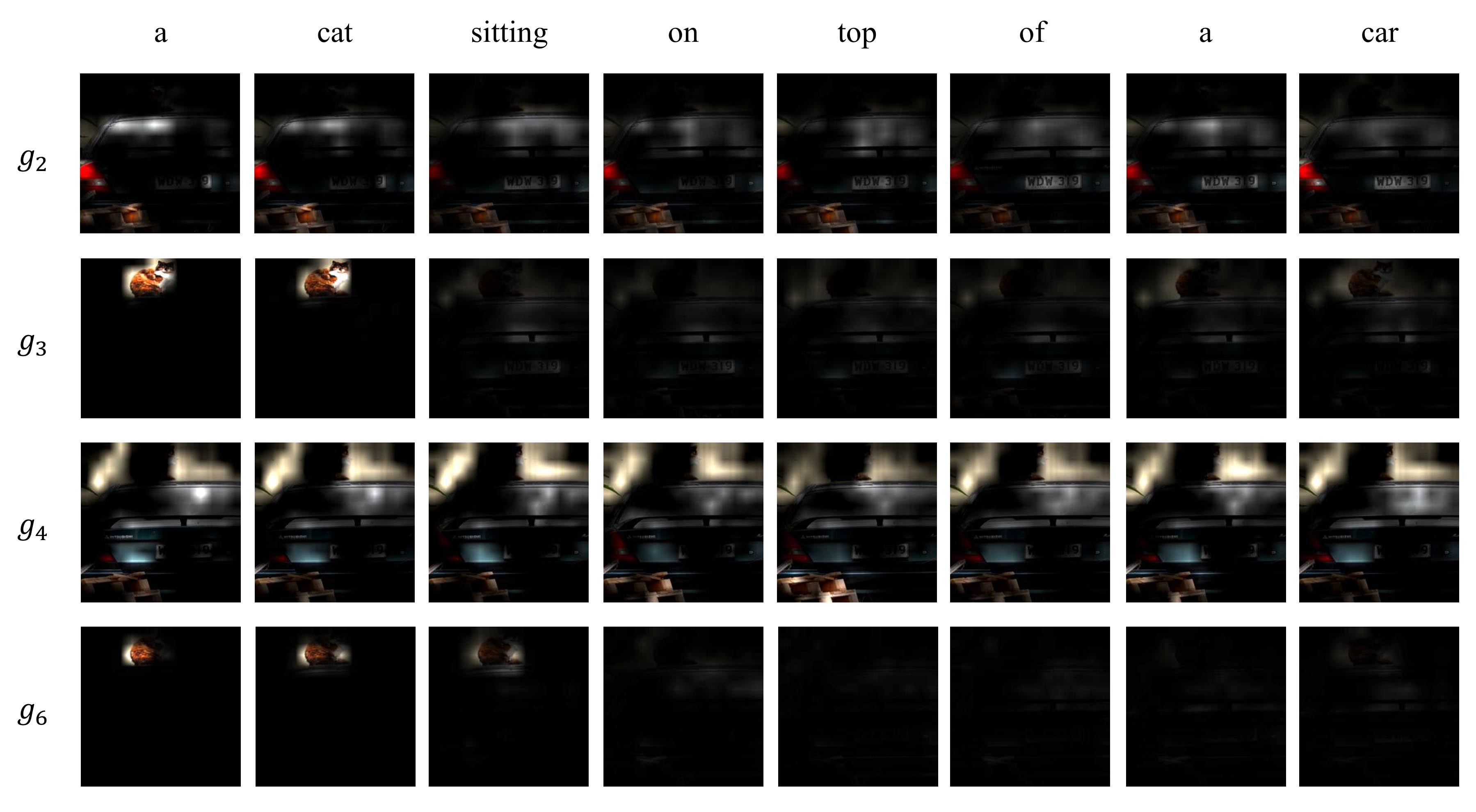}	\\
			
			\null & \\
			
			\gph{1.0}{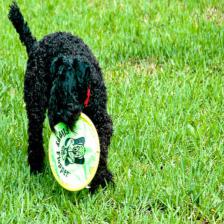}	&
			\gph{1.0}{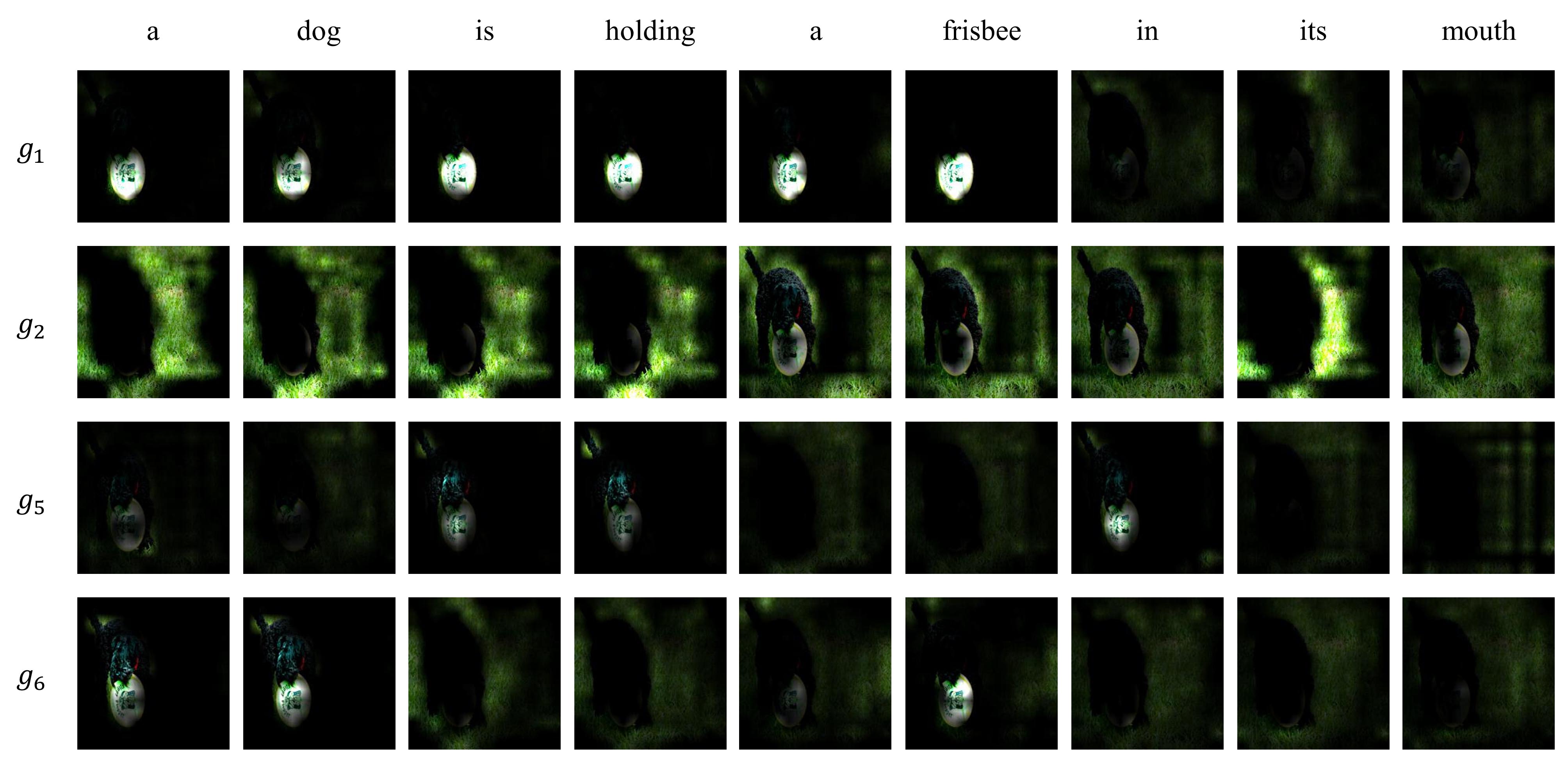}	\\
			
			\null & \\
			
			\gph{1.0}{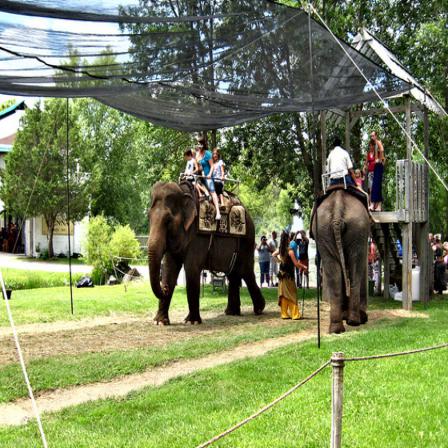}	&
			\gph{1.0}{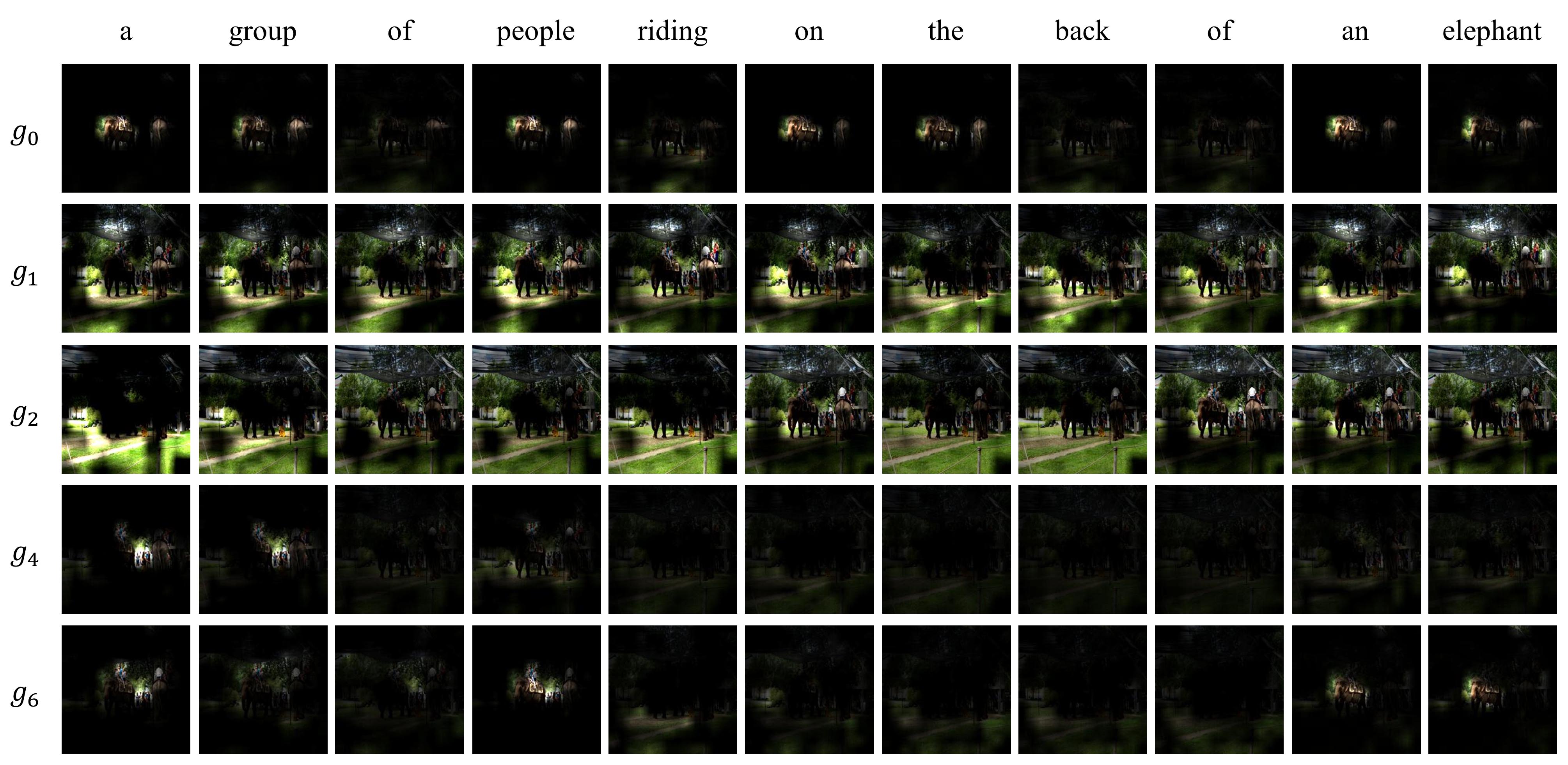}	\\
		
		\end{tabular}
	\end{adjustbox}
	\caption{Sample of the generated captions with the attention maps of different heads. It shows that COMIC has effectively delegated each attention head to different tasks.}
	\label{fig: Multi-attention 1}
\end{figure*}

\section{Quantitative Results}
\subsection{State-of-the-art comparison}
\label{subsubsec: SOTA comparison}

In this section, our COMIC model is a \Enc{} model with $8$ attention heads and tied feature map projection, the rest being identical to baseline.
To provide a fair comparison, we trained two sets of baseline models. The first set consists of the standard baseline named ``Baseline'' and ``Baseline-8'' with $8$ attention heads without feature map projection; while the second set consists of a pair of slim baseline models named ``Baseline-S'' where the parameter counts are reduced to match the COMIC models. ``Baseline-SC'' models have $n = k = 160$ and $m = 128$, and ``Baseline-SI'' models have $n = k = 80$ and $m = 64$.
We trained all the baselines and COMIC\,-$\,v$ models for $30$ epochs, where $v$ denotes the choice of base number. The base number of COMIC is chosen so that the number of tokens needed to encode a word token is $d = 2$ and $v^2 \geq |V_o|$. As MS-COCO word model has a vocabulary size of $V_o = 9,962$, a base number of $v = 128$ or $v = 256$ is sufficient to encode the entire $V_o$ while minimising the increase in sequence lengths. On the other hand, InstaPIC word model has $V_o = 25,598$, hence larger base numbers $v = 160$ and $256$ are used. We would like to note that our metric scores are obtained using a single model instead of an ensemble. 

Table \ref{table: SOTA Performance on MS-COCO}-\ref{table: SOTA Performance on InstaPIC} show the metric scores achieved by our baselines, COMIC and SOTA methods. On both datasets, our COMIC models managed to perform on par with the baselines even having much lower parameter count and vocabulary size. For example, on the MS-COCO dataset, the loss in performance of COMIC-256 is merely $0.45\%$ on CIDEr when compared to the baselines, despite with only $33\%$ of the parameters and a vocabulary size of $258$ ($39\times$ reduction). On the InstaPIC dataset, the complexity reduction is even more drastic. Despite having much lesser parameters ($16.7\%$ of baseline) and vocabulary size of only $258$ ($99\times$ reduction), COMIC-256 still manages to perform on par with baseline models and even outperforms it on certain metrics. When compared to the slim baseline models with comparable parameter counts, our COMIC models again have better performance. This shows the effectiveness of the proposed methods in reducing the model complexity and at the same time minimising its impact on overall performance across five different evaluation metrics. Our COMIC models also compare favourably to SOTA approaches, losing moderately to attribute-based approaches in the MS-COCO dataset, and only to the latest CSMN \cite{chunseong2017attend} and AACL \cite{dai2017contrastive} approaches in the InstaPIC dataset, despite operating on a much condensed vocabulary size.

As a summary, although there is a slight performance drop in some of the metric scores when comparing COMIC against the baselines in Table \ref{table: SOTA Performance on MS-COCO}-\ref{table: SOTA Performance on InstaPIC}, this performance degradation is an expected trade off of parameter reduction and we believe the results are still comparable. Then, when compared to the SOTA methods (with the exception of ACVT \cite{wu2016value} which implemented an attribute dictionary), we showed that the performance of our proposed model in overall is very competitive in both of the datasets. In particular, those that have similar architectures (\ie Soft and Hard Attention \cite{xu2015show}, and Review Net \cite{yang2016review}) as to our proposed work.

\subsection{Uniqueness and length of captions:}
\label{subsubsec: Captions stats}

It has been pointed out that multimodal RNN-based approaches tend to reconstruct previously seen captions \cite{devlin2015language}. Hence, we compare our model with baselines in terms of the uniqueness and length of the generated captions in Table \ref{table: Caption statistics}. A caption is considered to be unique if it's not seen in the pre-processed training corpus. From the results, we can see that although COMIC uses an encoded vocabulary, it still managed to generate considerably more unseen (unique) captions compared to the baselines. The average length of captions generated by the COMIC is also longer compared to the baselines. 

The trend is due to the decoding noise introduced by the \Enc{}. In other words, in addition to the long-term dependencies between words, successful generation of captions relies strongly on accurately modelling the short-term dependencies between tokens. This has increased the difficulty along with exposure bias for the increased uniqueness of captions generated by our models, as well as the increased length of captions.

\section{Qualitative Results}
\label{sec: Qualitative Results}

In this section, we provide some examples of the generated captions from our model in Fig. \ref{fig: captions MSCOCO}-\ref{fig: captions InstaPIC} for both of the MS-COCO and InstaPIC datasets\footnote{More results can be found in our supplementary material}. We can see that the captions generated by COMIC-256 are grammatically correct and are not affected by the vocabulary encoding scheme. In many cases, COMIC-256 even managed to provide finer details when describing the images compared to the baseline. For instance in the first image of Fig. \ref{fig: captions MSCOCO}, our model properly describes the image content {\it ``a man standing next to a zebra in a field"}, while the baseline model only able to generate {\it ``a man is standing next to a zebra"}.

To better understand our model, Fig. \ref{fig: Multi-attention 1} visualises the multi-head attention maps for different words in the generated caption. Going through each of the attention maps, we can see that our proposed model effectively delegates each attention head to different locations. In other words, each head learn to focus on subjects, objects or background separately. For example in the first image, we can visualise that the 3rd head ($g_2$) generally attends to the car. Meanwhile, the 4th head ($g_3$) is focused on the cat at the beginning, and fades out when the model moves to the other words. The 5th head ($g_4$) attends to the space around the roof of the car, aiding in predicting ``on top of''. Finally, the 7th head ($g_6$) attends to the boundary between the cat and the car while the model is predicting the word ``sitting''. The second image shows similar task assignments. In the third image that has multiple subjects, we can see that each head can separately attend to the background, elephant and the people sitting on top.


\section{Conclusion}
\label{sec: Conclusion}

This paper studied image captioning problem from a new perspective where it presented COMIC - a compact image captioning model with attention module. Experiments were conducted in the MS-COCO and InstaPIC-1.1M datasets, and the results showed that COMIC overall performance was not affected despite has a reduction of 33$\times$-99$\times$ in the vocabulary size. In future work, we would like to investigate the impact of different encoders (\ie CNN models) such as MobileNets \cite{howard2017mobilenets} on the overall performance and to train the \Enc{} models in a greedy decoding setting using reinforcement learning methods, such as Policy Gradient \cite{liu2017policy} to avoid the ``exposure bias'' problem.

\section*{Acknowledgement}
We gratefully acknowledge the support of NVIDIA Corporation with the donation of Titan Xp GPU used for this research.

\ifCLASSOPTIONcaptionsoff
\newpage
\fi
\balance

\bibliographystyle{IEEEtran}
\bibliography{references}


\clearpage
\onecolumn

\section*{Appendix}
In this supplementary material we provide additional visualisations of the attention heads in our Compact Image Captioning (COMIC) model on MS-COCO (in Sec. \ref{subsubsec: Attention Maps (MS-COCO)}) and InstaPIC-1.1M (in Sec. \ref{subsubsec: Attention Maps (InstaPIC)}) datasets. Furthermore, we also show some randomly sampled images with qualitative results in Sec. \ref{subsec: Generated Captions}.

\section{Multi-head Attention Maps}
\label{subsec: Multi-head Attention Maps}

In Fig. \ref{fig: Appendix Multi-attention 1} to \ref{fig: Appendix Multi-attention 6}, the attention maps of different heads are denoted by $g_a$ where $a = [0, 7]$. Attention maps with the most activity are selected for better visualisations. Going through each of the attention maps, we can see that the models have effectively learned how to delegate each attention head to different tasks. In other words, each head has learned to focus on subjects, objects or background separately.


\subsection{MS-COCO Dataset}
\label{subsubsec: Attention Maps (MS-COCO)}

\begin{figure*}[th]
	\small
	\centering
	\begin{tabular}{c}
		
		\gph{1.0}{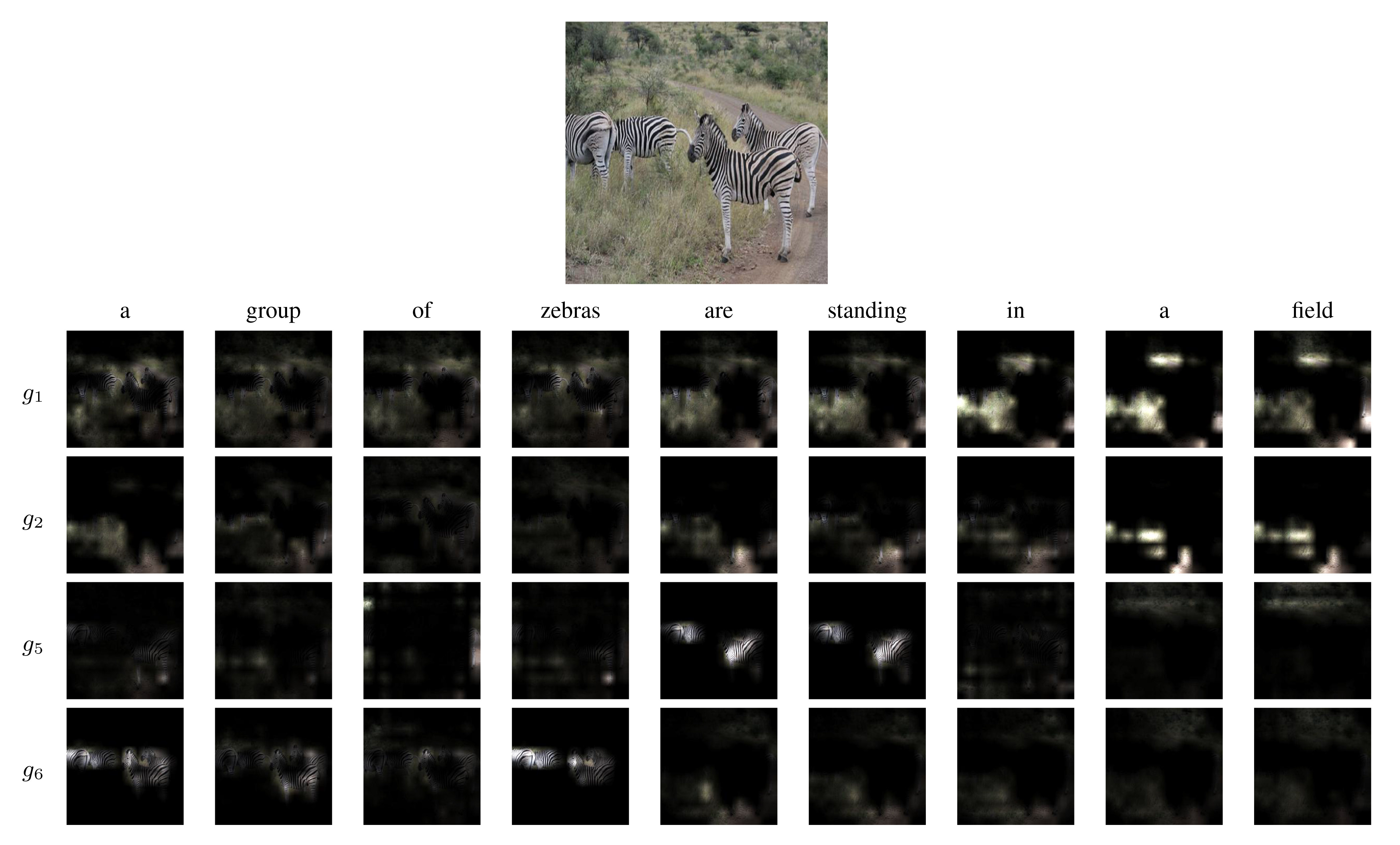}
		\\
		
	\end{tabular}
	\caption{Generated captions using greedy decoding by COMIC-256 on the MS-COCO dataset and the attention maps of different heads}
	\label{fig: Appendix Multi-attention 1}
\end{figure*}

{\bf Fig. \ref{fig: Appendix Multi-attention 1}}: We can see that both $g_1$ and $g_2$ attend to the surroundings of the zebras. While both $g_5$ and $g_6$ attend to the group of zebras, they provide different context to the language model as they switch on alternately of each other. $g_6$ likely provides context for the noun ``zebra'' while $g_5$ provides context for the verb ``standing''. \\

\begin{figure*}[th]
	\small
	\centering
	\begin{tabular}{c}
		
		\gph{1.0}{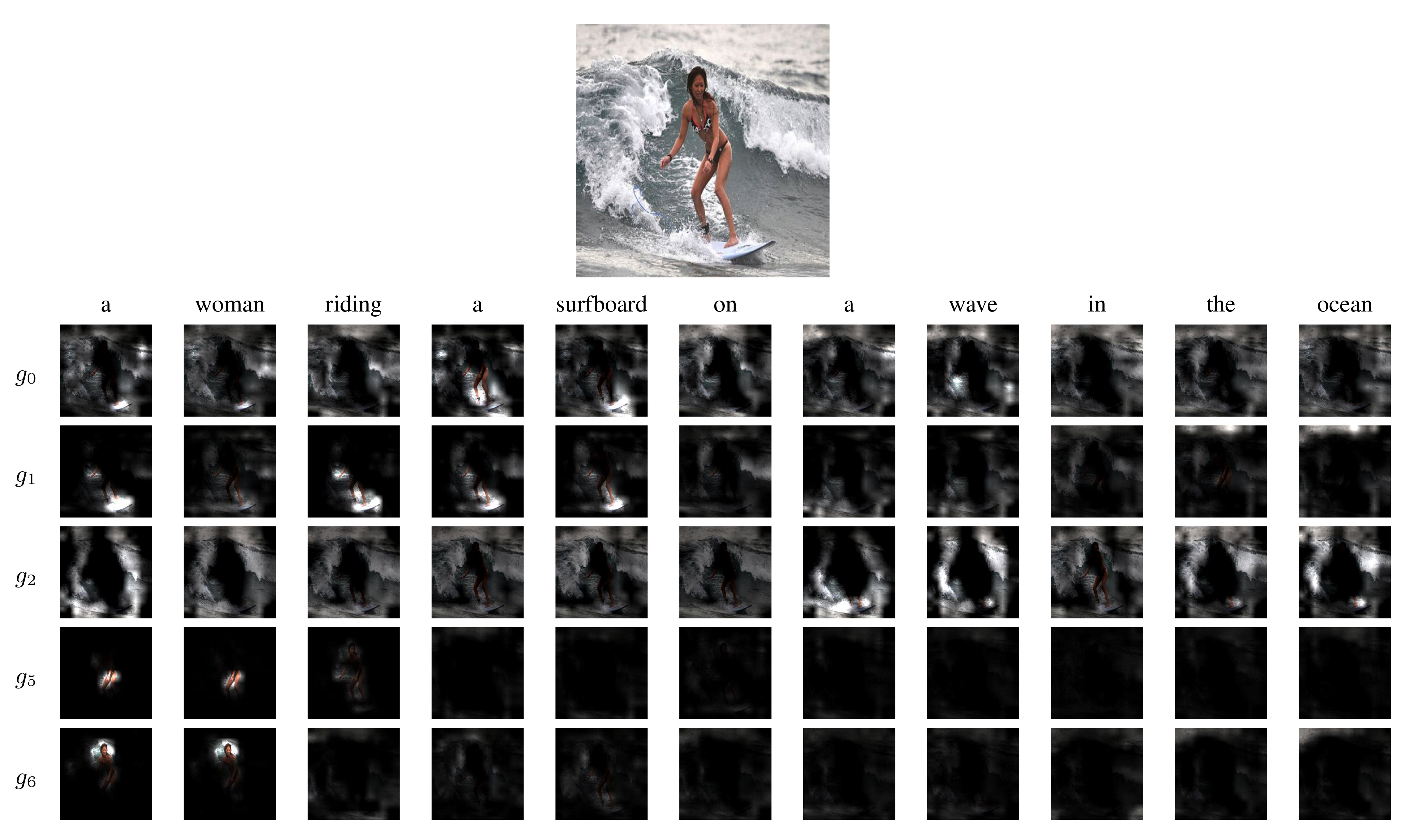}
		
		\\ \\ \\
		
		\gph{1.0}{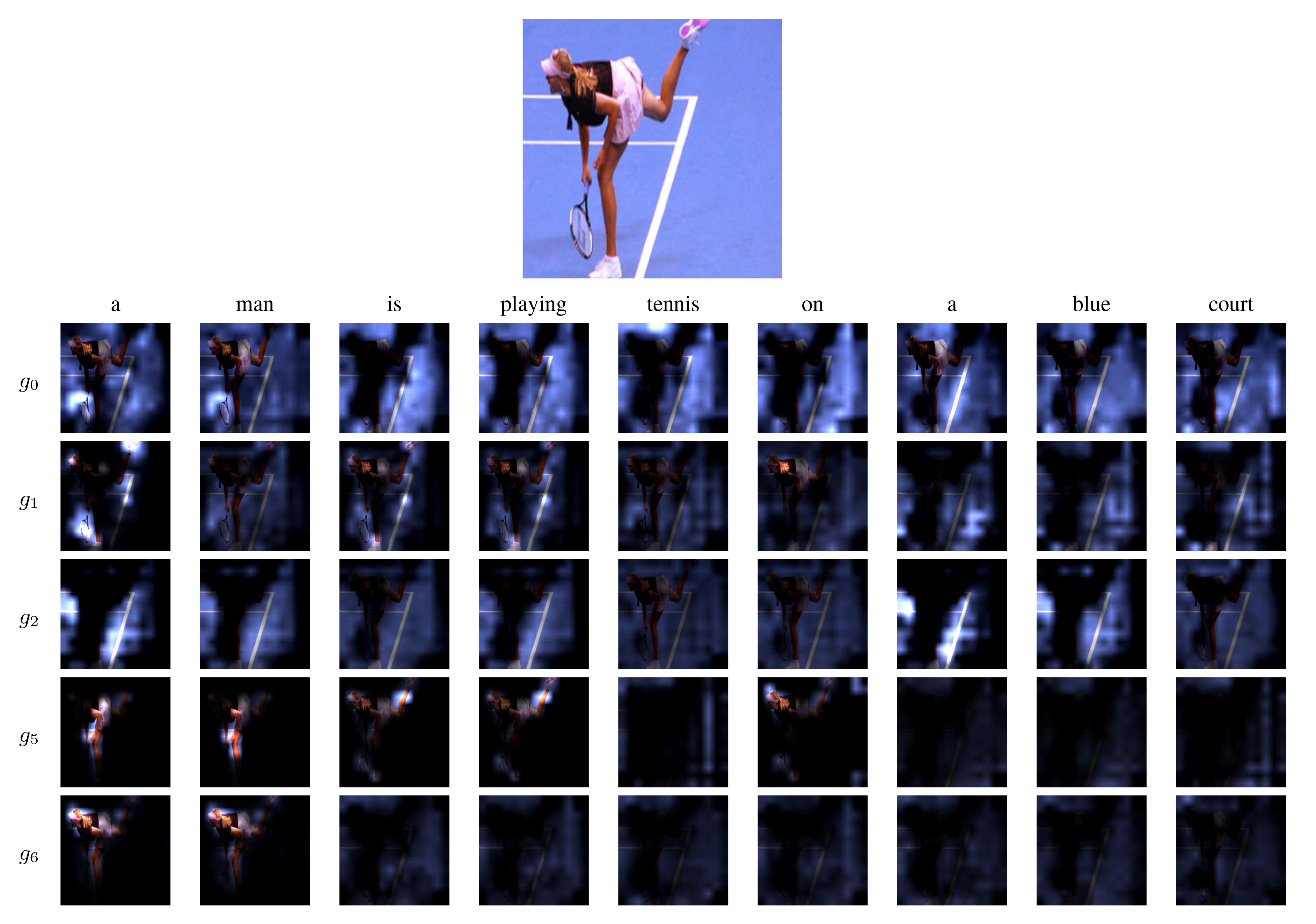}
		\\
		
	\end{tabular}
	\caption{More examples on MS-COCO dataset: Generated captions and the attention maps of different heads}
	\label{fig: Appendix Multi-attention 2}
\end{figure*}

{\bf Fig. \ref{fig: Appendix Multi-attention 2} (top)}: Here, we can see that both $g_0$ and $g_1$ attend to the surfboard, but $g_0$ also attends to the surrounding ocean which might provide the general context. In contrast, $g_1$ is strongly focused on the surfboard. $g_2$ attends to the waves. Lastly, both $g_5$ and $g_6$ attend to the subject with $g_5$ focusing on the lower body and $g_6$ focuses on the head and torso.

{\bf Fig. \ref{fig: Appendix Multi-attention 2} (bottom)}: Although the model {\it misidentifies} the player as male, the attention is focused on the correct regions. $g_0$ attends to the player and the court, that provide the general context. $g_1$ attends to the cap, racquet, shoes and clothing. This provides the cue on the type of sports, which our model predicted correctly. $g_2$ again attends to the background, in particular the court lines. Similar to the surfing example above, both $g_5$ and $g_6$ attend to the subject with $g_5$ focuses on the lower body and $g_6$ focuses on the head and torso.

\clearpage
\subsection{InstaPIC-1.1M Dataset}
\label{subsubsec: Attention Maps (InstaPIC)}

\begin{figure*}[th]
	\small
	\centering
	
	\gph{1.0}{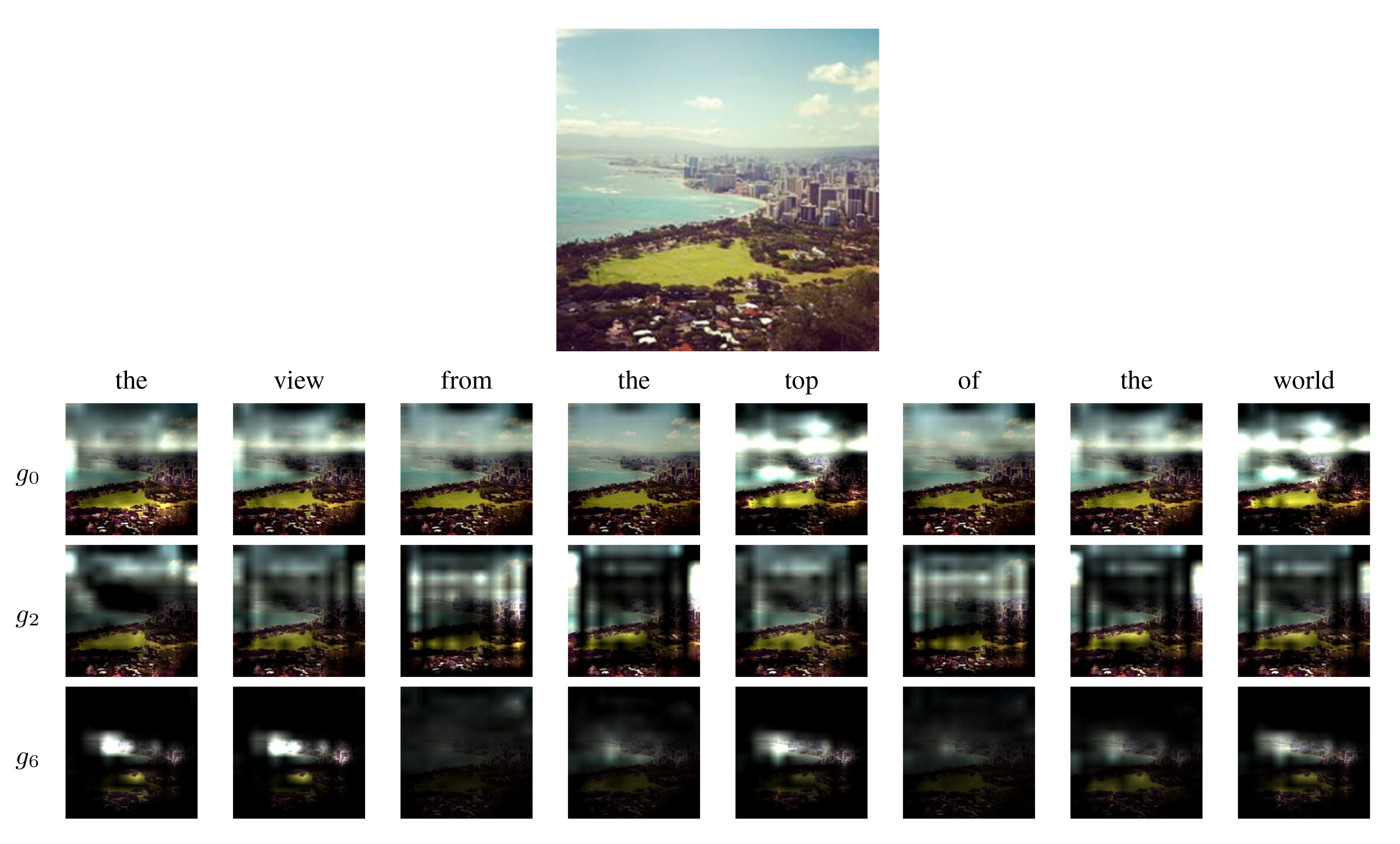}
	
	\caption{Generated captions using greedy decoding by COMIC-256 on the InstaPIC-1.1M dataset and the attention maps of different heads}
	\label{fig: Appendix Multi-attention 4}
\end{figure*}

{\bf Fig. \ref{fig: Appendix Multi-attention 4}}: We can see that $g_0$ attends mainly to the sky region especially when the model is predicting ``top'' and ``world''. Basically, $g_2$ attends to the entire image, which provide the general context. Lastly, $g_6$ attends to both the foreground and faraway regions, which provide cues that the image is a bird's-eye view of the bay region. \\

\begin{figure*}[th]
	\small
	\centering
	\begin{tabular}{c}
		
		\gph{1.0}{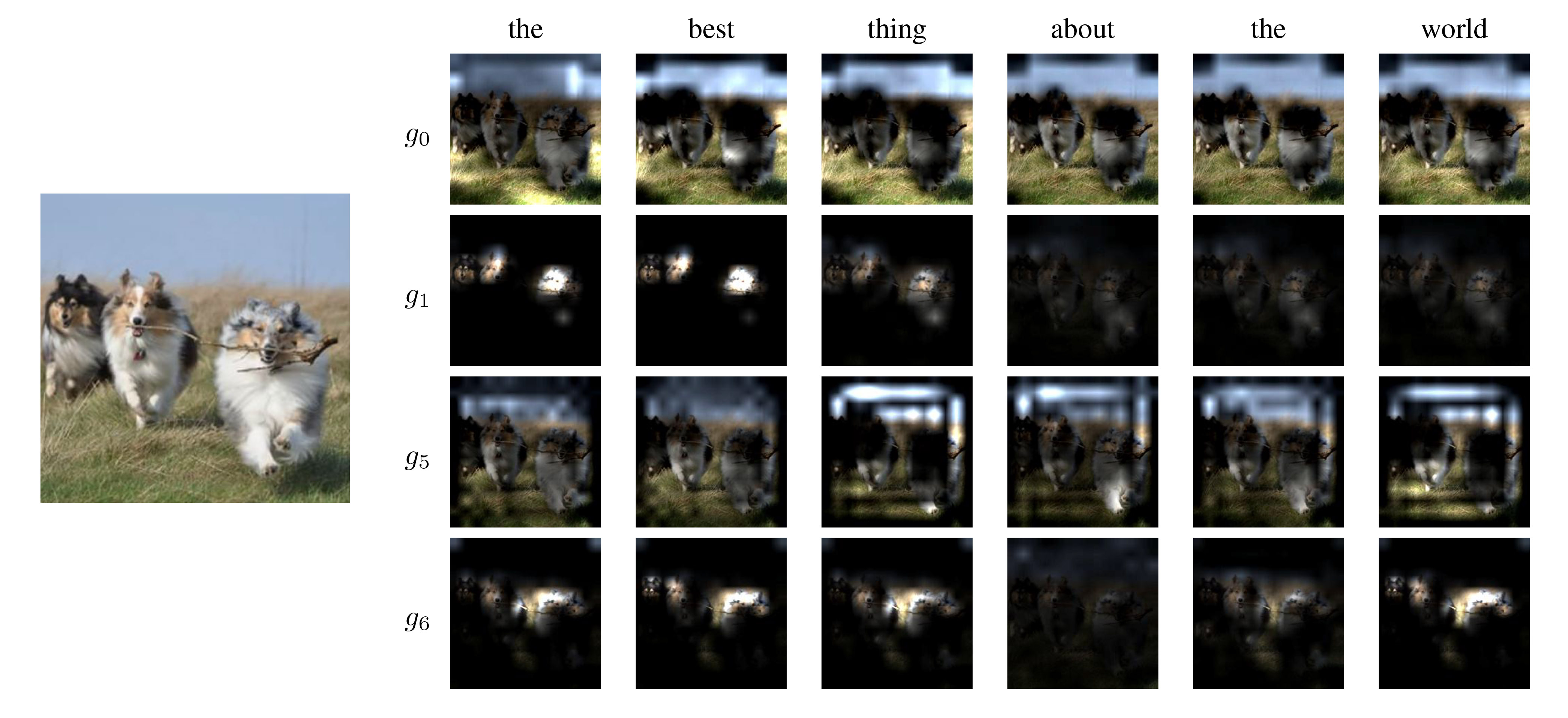}
		
		\\ \\ \\
		
		\gph{1.0}{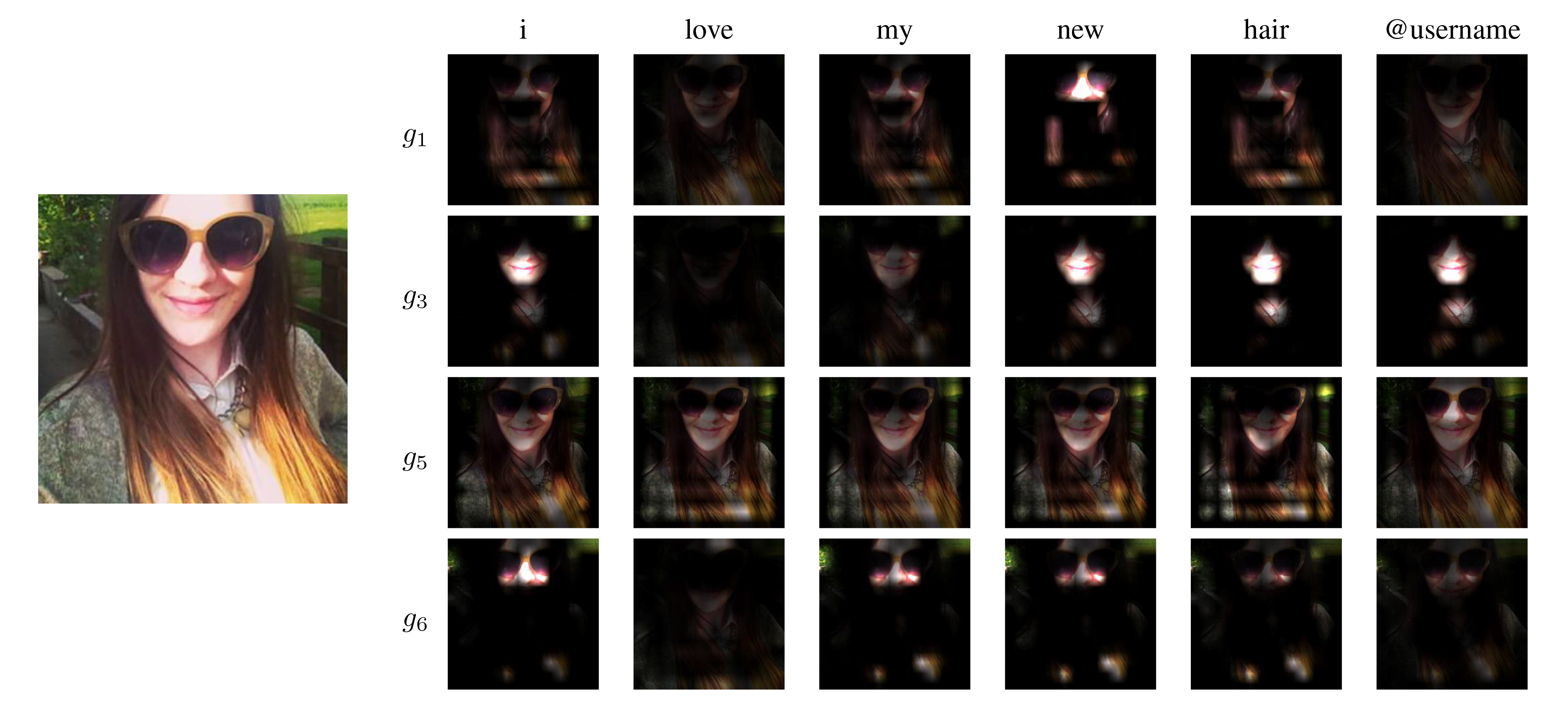}
		\\
		
	\end{tabular}
	\caption{More examples on InstaPIC-1.1M dataset: Generated captions and the attention maps of different heads}
	\label{fig: Appendix Multi-attention 5}
\end{figure*}

{\bf Fig. \ref{fig: Appendix Multi-attention 5} (Top)}: It can be seen that $g_0$ attends to the background. $g_5$ attends to basically the entire image, which provide the general context. Both $g_1$ and $g_6$ attends to the dog, with $g_1$ is more focused than $g_6$.

{\bf Fig. \ref{fig: Appendix Multi-attention 5} (Bottom)}: We can see that $g_1$ attends to the hair and face of the subject, while both $g_3$ and $g_6$ attend strongly to the facial regions. $g_5$ attends to the entire image. \\

\begin{figure*}[th]
	\small
	\centering
	\begin{tabular}{c}
		
		\gph{1.0}{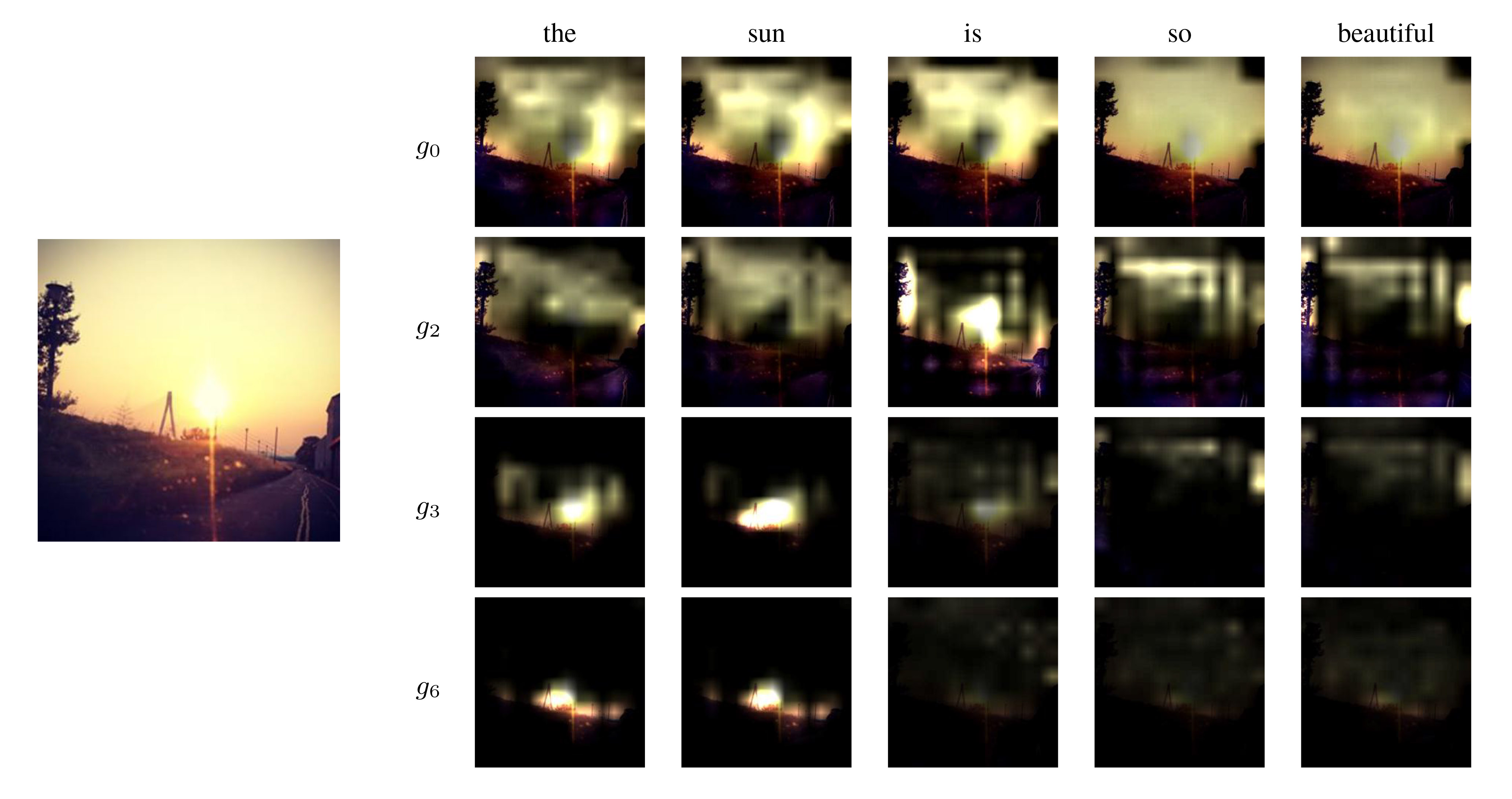}
		
		\\ \\ \\
		
		\gph{1.0}{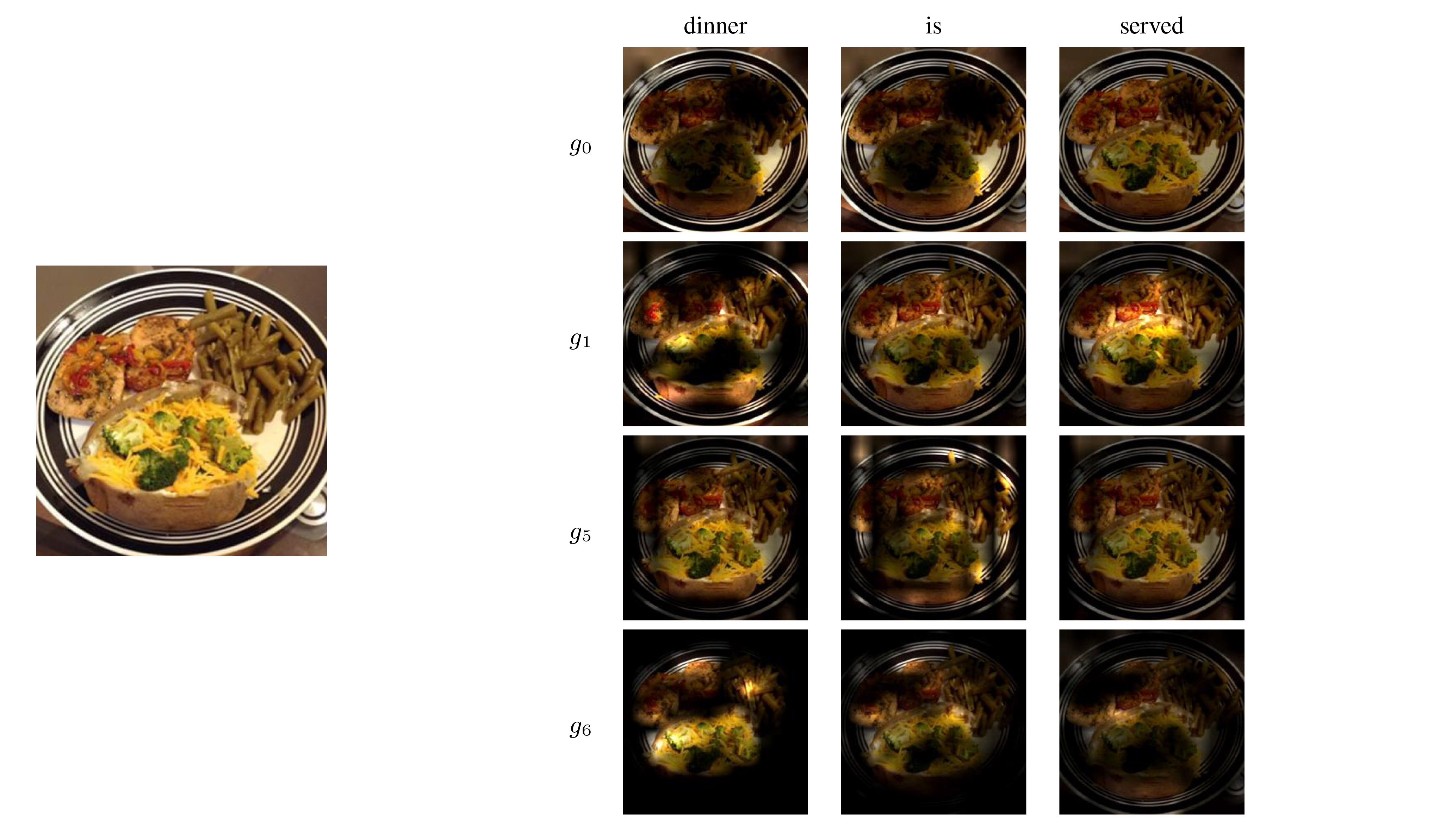}
		\\
		
	\end{tabular}
	\caption{More examples on InstaPIC-1.1M dataset: Generated captions and the attention maps of different heads}
	\label{fig: Appendix Multi-attention 6}
\end{figure*}

{\bf Fig. \ref{fig: Appendix Multi-attention 6} (Top)}: We can clearly observe that $g_0$ attends to the sky regions, while $g_2$ attends to the tree, road and sun. Both $g_3$ and $g_6$ attend to the sun, with $g_3$ being more focused than $g_6$.

{\bf Fig. \ref{fig: Appendix Multi-attention 6} (Bottom)}: Here, we can see that $g_0$ mainly attends to the plate, while $g_5$ attends to the food as a whole. Both $g_1$ and $g_6$ attend to different regions or food items on the plate.

\bigskip
\bigskip

\clearpage
\section{Generated Captions}
\label{subsec: Generated Captions}

\begin{figure*}[h]
	\footnotesize
	\centering
	
	\gph{0.82}{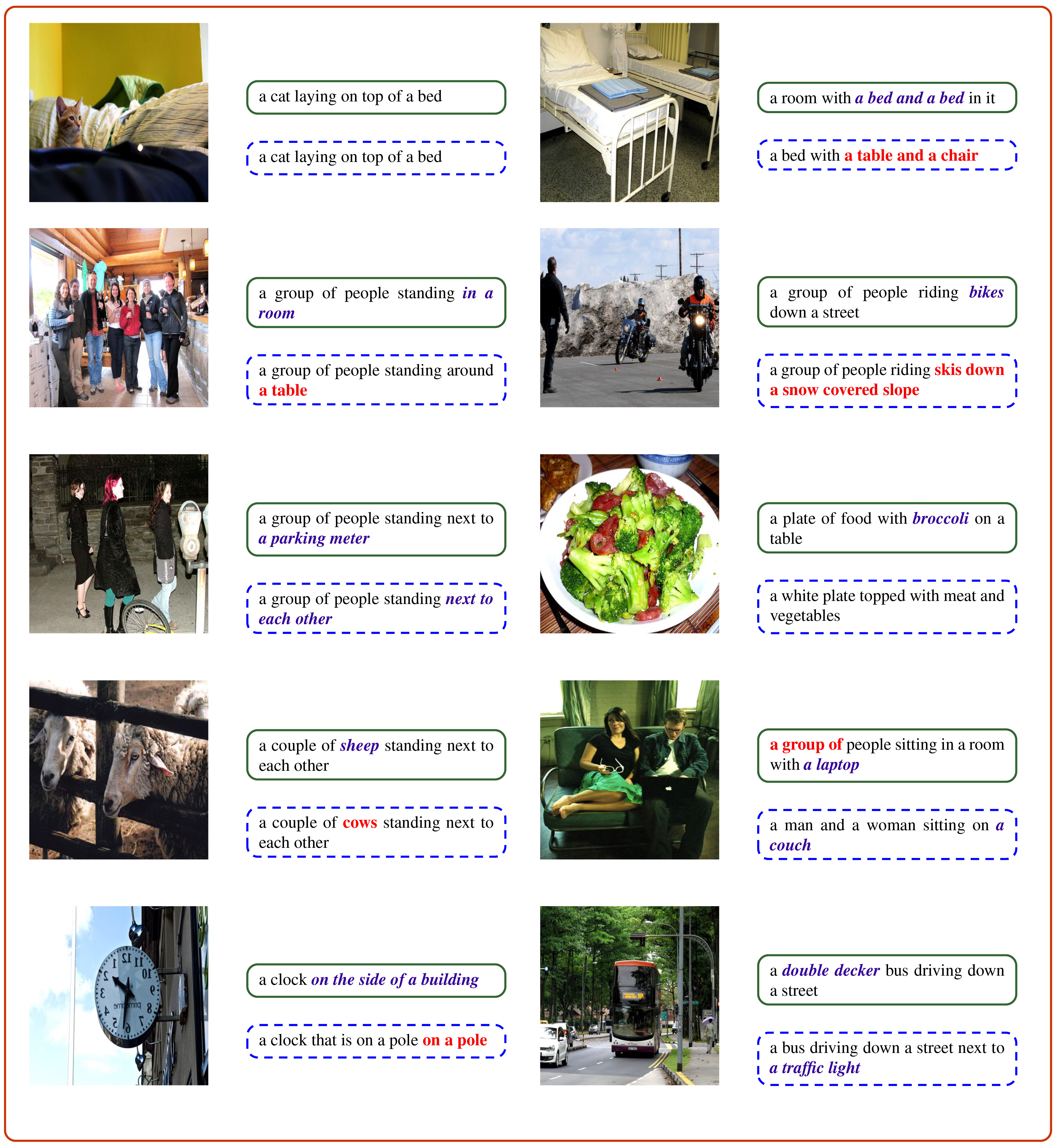}
	
	\caption{Captions generated by COMIC-256 (solid green line) and baseline (dashed blue line) on MS-COCO dataset. Accurate descriptions are indicated by \correct{blue with bold and italic text}, inaccurate descriptions are indicated by \wrong{red with bold text}. Best viewed in colour}
	\label{fig: Appendix MSCOCO 1}
\end{figure*}

\begin{figure*}[th]
	\footnotesize
	\centering
	
	\gph{1.0}{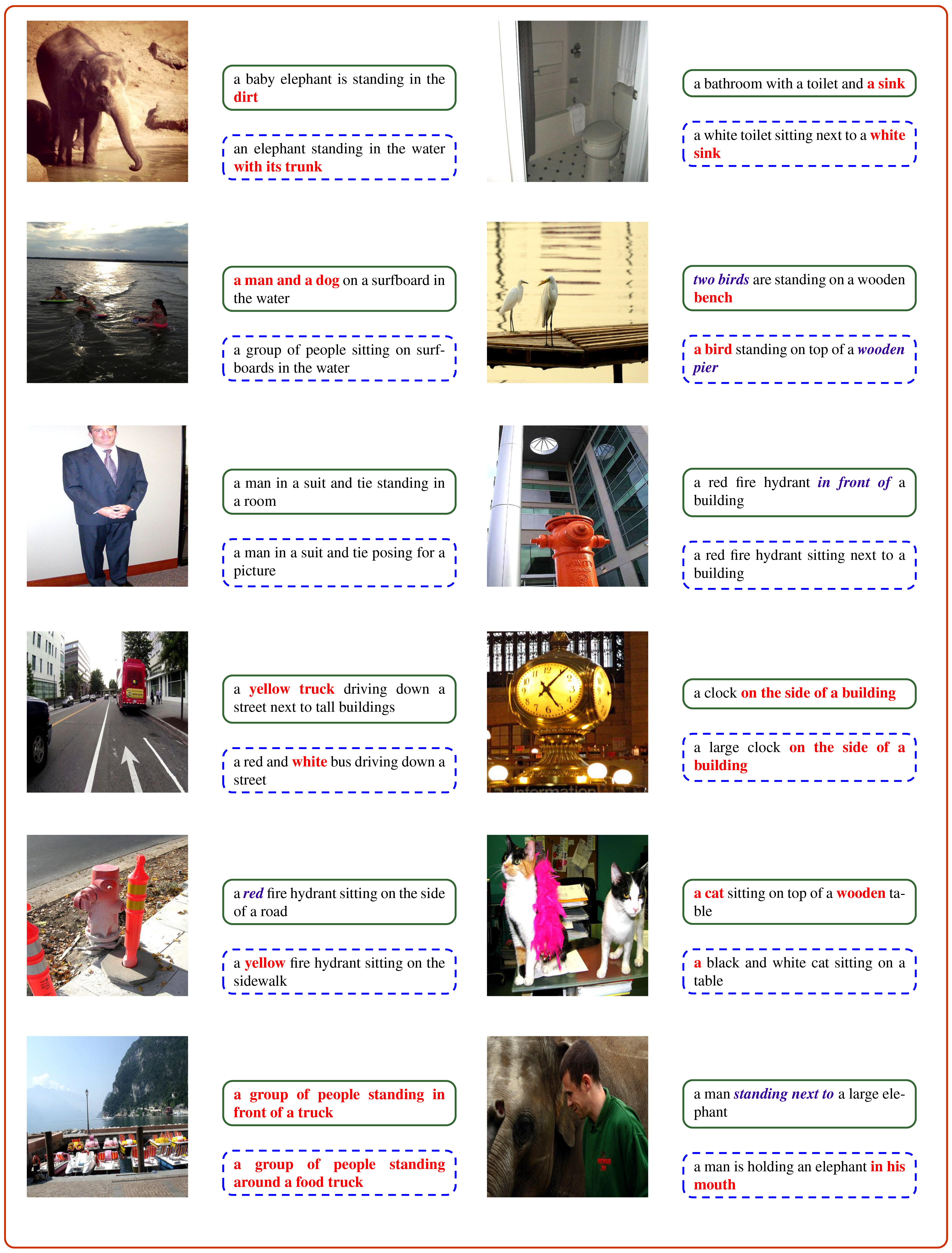}
	
	\caption{Captions generated by COMIC-256 (solid green line) and baseline (dashed blue line) on MS-COCO dataset. Accurate descriptions are indicated by \correct{blue with bold and italic text}, inaccurate descriptions are indicated by \wrong{red with bold text}. Best viewed in colour}
	\label{fig: Appendix MSCOCO 2}
\end{figure*}

\begin{figure*}[th]
	\footnotesize
	\centering
	
	\gph{0.92}{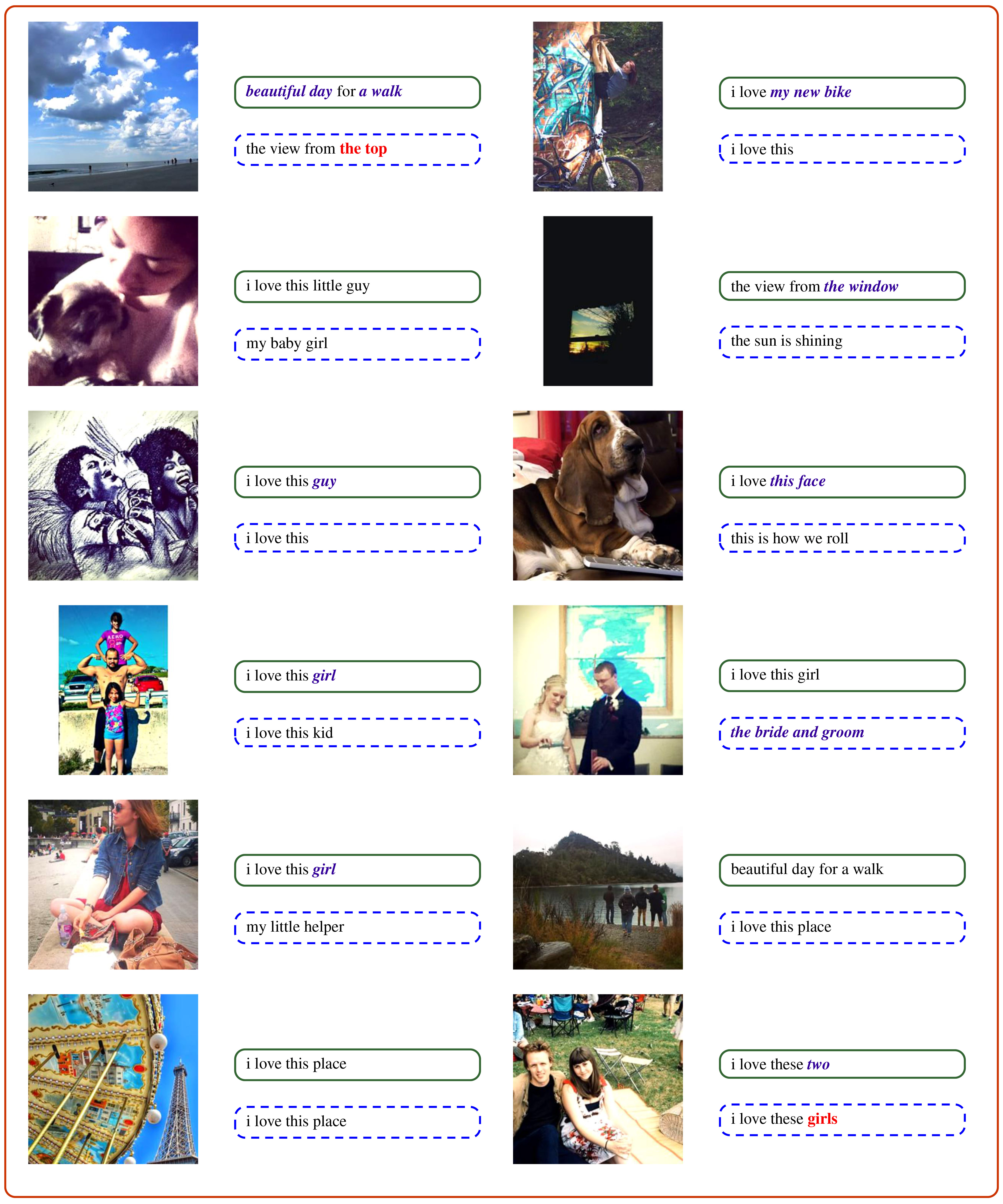}
	
	\caption{Captions generated by COMIC-256 (solid green line) and baseline (dashed blue line) on InstaPIC-1.1M dataset. Accurate descriptions are indicated by \correct{blue with bold and italic text}, inaccurate descriptions are indicated by \wrong{red with bold text}. Best viewed in colour}
	\label{fig: Appendix Insta 1}
\end{figure*}

\begin{figure*}[th]
	\footnotesize
	\centering
	
	\gph{1.0}{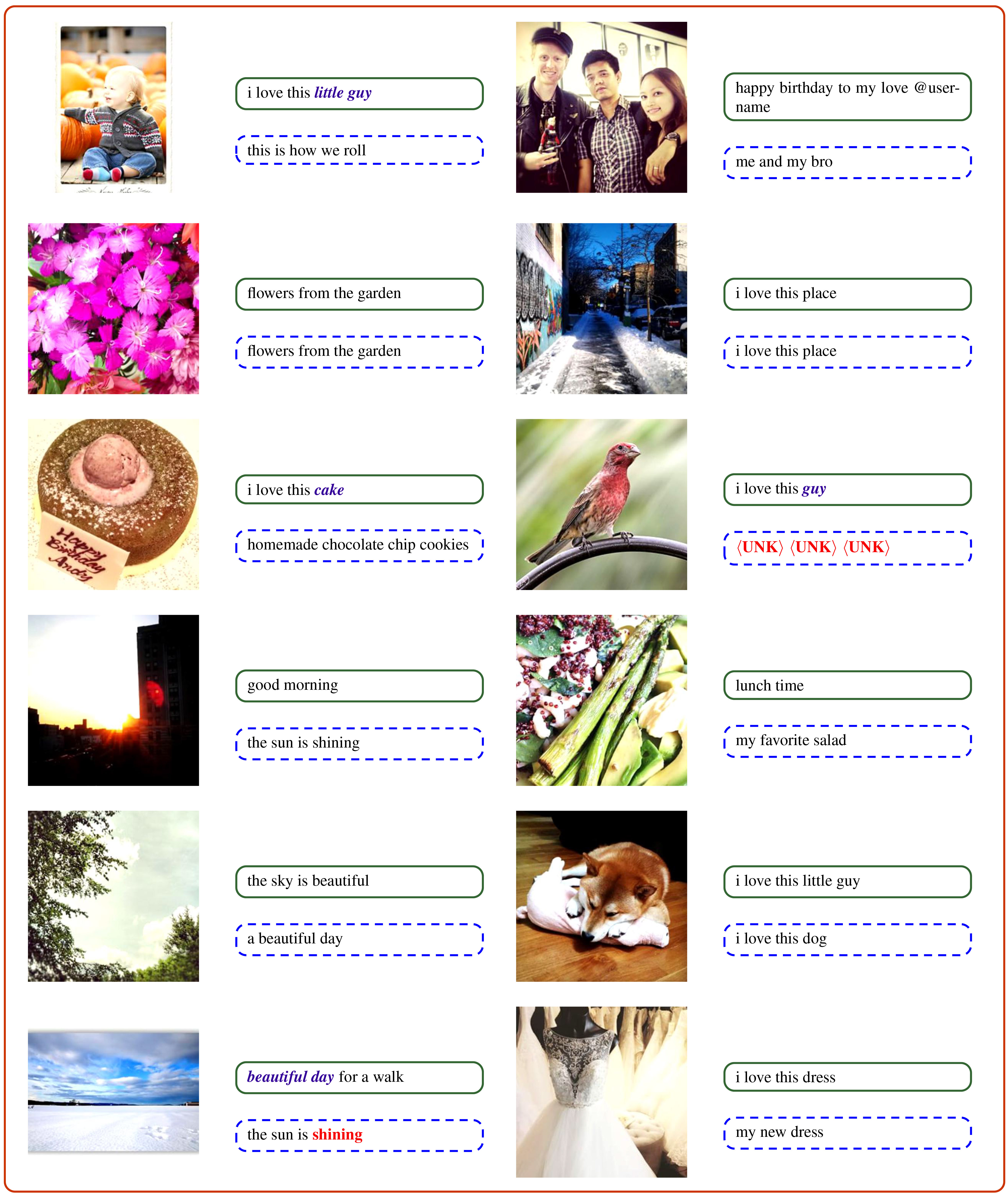}
	
	\caption{Captions generated by COMIC-256 (solid green line) and baseline (dashed blue line) on InstaPIC-1.1M dataset. Accurate descriptions are indicated by \correct{blue with bold and italic text}, inaccurate descriptions are indicated by \wrong{red with bold text}. Best viewed in colour}
	\label{fig: Appendix Insta 2}
\end{figure*}

\begin{figure*}[th]
	\footnotesize
	\centering
	
	\gph{1.0}{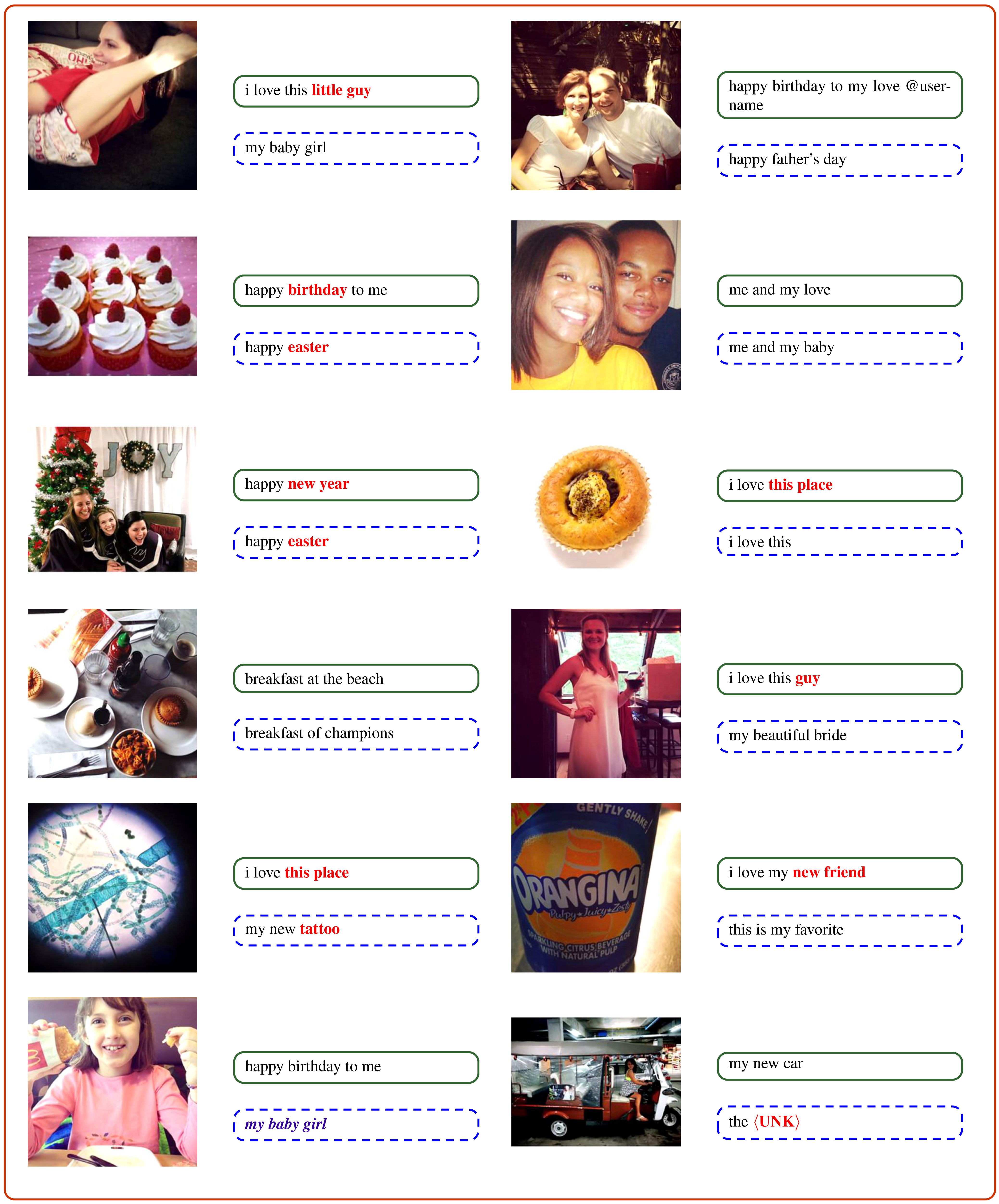}
	
	\caption{Captions generated by COMIC-256 (solid green line) and baseline (dashed blue line) on InstaPIC-1.1M dataset. Accurate descriptions are indicated by \correct{blue with bold and italic text}, inaccurate descriptions are indicated by \wrong{red with bold text}. Best viewed in colour}
	\label{fig: Appendix Insta 3}
\end{figure*}

For the generated captions, we provide results from both our COMIC-256 model and the baseline word model in Figure \ref{fig: Appendix MSCOCO 1} to \ref{fig: Appendix Insta 3}. Captions inside the {\color{OliveGreen} solid green box} are generated by COMIC-256 model, and captions inside the {\color{DarkBlue} dashed blue box} are generated by the baseline method. We can see that for most images, our proposed method matches or in some cases outperforms the baseline method. For instance, we can see that the captions generated by COMIC-256 model are grammatically correct and this shows that it does not affected by the vocabulary encoding scheme. In some cases, COMIC-256 model managed to provide finer details when describing the images compared to the baseline. Finally, we demonstrate the ability of our proposed method to generate variable length captions.

We also explicitly chose some failure examples in which COMIC-256 model performs no better than baseline method in Figure \ref{fig: Appendix MSCOCO 2} for MS-COCO dataset and Figure \ref{fig: Appendix Insta 3} for InstaPIC-1.1M dataset. We can see that incorrect recognition of objects or missing main objects in the image is still the dominant cause of error.


\subsection{MS-COCO Dataset}
\label{subsubsec: Captions (MS-COCO)}

\clearpage
\subsection{InstaPIC-1.1M Dataset}
\label{subsubsec: Captions (InstaPIC)}

\end{document}